\documentclass[10pt,twocolumn,letterpaper]{article}

\usepackage{iccv}              

\definecolor{iccvblue}{rgb}{0.21,0.49,0.74}
\usepackage[pagebackref,breaklinks,colorlinks,allcolors=iccvblue]{hyperref}
\usepackage{xcolor}  
\usepackage{booktabs}
\usepackage{makecell} 
\usepackage{tabularx}
\usepackage{stmaryrd}
\usepackage{multirow}
\usepackage{graphicx}

\definecolor{improvegreen}{RGB}{0, 128, 0} 
\definecolor{improveblue}{RGB}{0, 0, 255} 
\def\paperID{4458} 
\def\confName{ICCV}
\def\confYear{2025}

\title{Can Atomic Step Decomposition Enhance the Self-structured Reasoning of Multimodal Large Models?}

\author{Kun Xiang$^{1}$\thanks{These authors contributed equally to this work.}, Zhili Liu$^{2*}$, Zihao Jiang$^{3*}$, Yunshuang Nie$^{1}$, Kaixin Cai$^{1}$, Yiyang Yin$^{1}$,
Runhui Huang$^{4}$, \\Haoxiang Fan$^{1}$, Hanhui Li$^{1}$, Weiran Huang$^{3}$, Yihan Zeng$^{5}$, Yu-Jie Yuan$^{5}$, Jianhua Han$^{5}$,\\ 
Lanqing Hong$^{5}$, Hang Xu$^{5}$, Xiaodan Liang$^{1}$\thanks{Corresponding author. Email: \texttt{xdliang328@gmail.com}}
\\
$^{1}$ Sun Yat-sen University 
$^{2}$ Hong Kong University of Science and Technology \\
$^{3}$ Shanghai Jiaotong University
$^{4}$ University of Hong Kong
$^{5}$ Huawei Noah's Ark Lab \\
}

\begin{document}
\maketitle
\begin{abstract}

In this paper, we address the challenging task of multimodal mathematical reasoning by incorporating the ability of "slow thinking" into multimodal large language models (MLLMs). Our core idea is that different levels of reasoning abilities can be combined dynamically to tackle questions with different complexity. To this end, we propose a paradigm of Self-structured Chain of Thought (SCoT), which is composed of minimal semantic atomic steps. Different from existing methods that rely on structured templates or free-form paradigms, our method can not only generate cognitive CoT structures for various complex tasks but also mitigates the phenomenon of overthinking. To introduce structured reasoning capabilities into visual understanding models, we further design a novel AtomThink framework with four key modules, including (i) a data engine to generate high-quality multimodal reasoning paths; (ii) a supervised fine-tuning process with serialized inference data;  (iii) a policy-guided multi-turn inference method; and (iv) an atomic capability metric to evaluate the single step utilization rate. We conduct extensive experiments to show that the proposed AtomThink significantly improves the performance of baseline MLLMs, achieving more than 10\% average accuracy gains on MathVista and MathVerse. Compared to state-of-the-art structured CoT approaches, our method not only achieves higher accuracy but also improves data utilization by 5 times and boosts inference efficiency by 85.3\%. Our code is now public available in https://github.com/Quinn777/AtomThink.

\end{abstract}    
\section{Introduction}
\label{sec:intro}


\begin{figure*}[t]
    \centering
\includegraphics[width=1.0\textwidth]{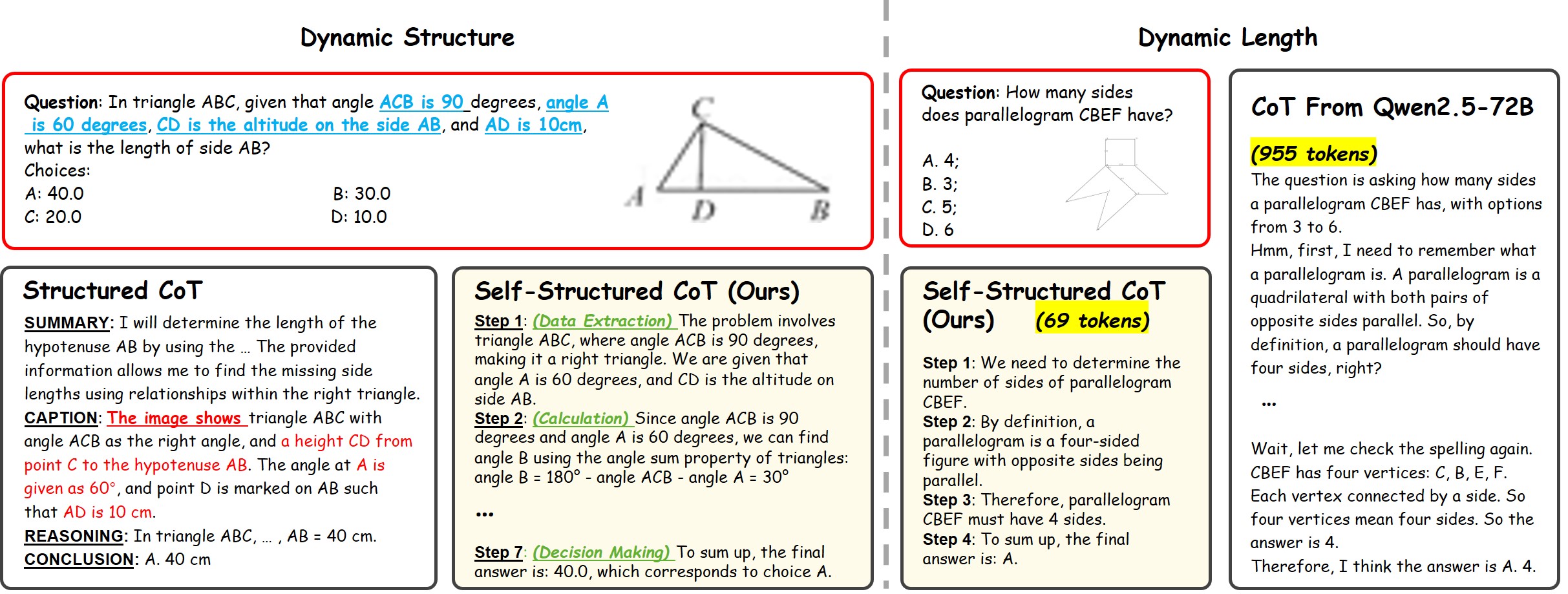} 
    \caption{Comparison with structured and unstructured reasoning models. We are capable of autonomously generating dynamic structures and lengths based on the type of problem. For text-dominant questions as shown on the left, we bypass image caption and directly extracted information from the question stem. For the low-difficulty problem on the right, we use fewer tokens compared to o1-like model.}
    \label{fig:intro}
\end{figure*}

Chain-of-Thought (CoT) reasoning~\cite{wei2022chain} has provided a novel scheme for Large Language Models (LLMs) to tackle complex reasoning tasks. By utilizing a small number of specially designed instructions, CoT enables LLMs to generate unstructured reasoning steps to enhance their performance on complex tasks. Moreover, the introduction of OpenAI's o1~\cite{openai2024_o1} then marks a substantial advancement in the ability of artificial intelligence systems to perform high-level reasoning. Unlike traditional models, o1 excels in solving complex problems by utilizing extended reasoning chains and adopting test-time scaling, i.e., ``\emph{slow thinking}''. 

More recently, numerous studies have ventured to hypothesize about their reasoning paradigms~\cite{yao2024tree,gao2024interpretable,qin2024o1,wang2024openr}. Certain endeavors, such as LLaVA-CoT~\cite{llavacot} and LlamaV-o1~\cite{llamavo1}, have posited structured methodologies employing fixed modules to drive reasoning. However, these methods require manually designed thinking templates, which limits the diversity of their reasoning behaviors in multimodal complex problems. In contrast to structured approaches, the advent of DeepSeek-R1~\cite{guo2025deepseek} has redirected attention towards unstructured reasoning. While unstructured chains of thought more closely mirror human cognitive patterns and exhibit superior generalization capabilities, recent studies~\cite{chen2024not, wang2025thoughts} have found that such slow-thinking models exhibit low efficiency in token utilization and demonstrate overthinking behavior when addressing simple problems. Example in Figure \ref{fig:intro} illustrates the challenges faced by both structured and unstructured CoTs. Therefore, we propose two clarifications: \textbf{1) Different types of problems may require distinct reasoning capabilities; 2) The complexity of reasoning should align with the difficulty of the problem.}



To dynamically generate appropriate reasoning structures for problems with diverse complexity, we introduce a novel paradigm of \textbf{Self-structured Chain-of-Thought (SCoT)}, which decomposes reasoning processes into minimal semantic atomic steps. To activate the model's self-structured reasoning abilities in multimodal tasks, we further develop a full-process slow-thinking framework called \textbf{AtomThink}. It comprises four key components, including a data engine and methods for atomic fine-tuning, policy search and atomic capability evaluation. To begin with, a data annotation engine with novel prompting and bad-case filtering strategies is used to create a novel multimodal long CoT dataset. We propose a dataset called AMATH, including 20k high-level mathematical problems with 124k atomic-granularity step annotations. Secondly, our atomic step finetuning strategy applies step-level masking to the training set, forcing our models to learn individual inference steps. During the inference phase, the model is not only capable of spontaneously generating CoT in quick mode, but also can be continuously improved with process supervision models and step search mechanisms. Lastly, we propose an atomic capability evaluation metric based on reasoning behavior clustering and step utilization calculation to evaluate the model's performance in utilizing individual atomic steps for reasoning. 

To validate the effectiveness of our method, we conduct extensive experiments on public datasets. We improve the accuracy of baseline on MathVista, MathVerse and MathVision by \textbf{10.9\%, 10.2\% and 7.2\%}, respectively. Furthermore, our model achieves data utilization at \textbf{500\%} of LLaVA-CoT while delivering superior performance, and improves inference efficiency by more than \textbf{80\%}. Aiming to advance the development of multimodal high-level reasoning, we also share extended analysis in diverse reasoning ability required by visual understanding models.

In summary, our primary contributions are as follows:
\begin{itemize}
    \item We introduce the Self-structured Chain-of-Thought to decompose the reasoning process into atomic steps. It eliminates the need for constructing structured thought templates and achieves significant improvements in both data utilization and inference efficiency.
  \item A comprehensive framework including modules for data annotation, atomic fine-tuning, multi-turn inference and capability evaluation, is designed to improve the reasoning ability of MLLMs.
  \item We validated the effectiveness of our approach on three high-level reasoning benchmarks and across various scales of MLLMs. Additionally, we present an analysis of the distribution of comprehension capabilities in MLLMs.
\end{itemize}

\section{Related Work}
\label{sec:formatting}
\paragraph{Chain of Thought in Multimodal Reasoning Tasks}
Complex reasoning tasks such as mathematical computation have long been challenging for MLLMs~\cite{yin2023survey,liu2023mathematical}. Some prior work has approached this issue from the perspective of prompt engineering, encouraging models to generate CoT, which is widely believed to enhance reasoning abilities \cite{wei2022chain,wang2022self}. They carefully modify the input distribution to generate unstructured reasoning path without finetuning parameters. Recently, OpenAI o1 and DeepSeek R1 have demonstrated the scalability of unstructured CoT by guiding models to autonomously learn reasoning patterns through Reinforcement Learning. However, reasoning models still suffer from issues of overthinking and excessive computational consumption. Other studies have guided multimodal models to generate structured CoT by providing manually designed templates \cite{llavacot,llamavo1}. Although these models incorporate visual semantic information into the reasoning process, their fixed steps constrain the diversity of reasoning actions, limiting their generalization ability on complex problems.

\paragraph{Long CoT Annotation for Multimodal Data}
The introduction of slow thinking relies heavily on the availability of high-quality step-level annotations. Lightman et al. \cite{lightmanlet} constructed a process supervision dataset composed of extensive human annotations, which has been widely used for mathematical reasoning. Recent advancements have focused on automating the data acquisition process, allowing models to generate their own CoTs. Techniques like Quiet-STaR \cite{zelikman2024quiet} have demonstrated how self-generated reasoning can enhance model performance without requiring manual labels. Moreover, some methods based on Monte Carlo estimation have automated the process of data collection, but they also introduce additional computational cost \cite{Math-shepherd,OmegaPRM}. In multimodal domain, MAVIS \cite{zhang2024mavis}, a dataset consisting of 834k visual math problems annotated with short CoT, has been proposed. Other studies have distilled reasoning processes from short answers \cite{zhang2024improve}. However, these machine-generated annotations are often too brief and challenging to segment semantically.

\begin{figure*}[t]
    \centering
    \includegraphics[width=\textwidth]{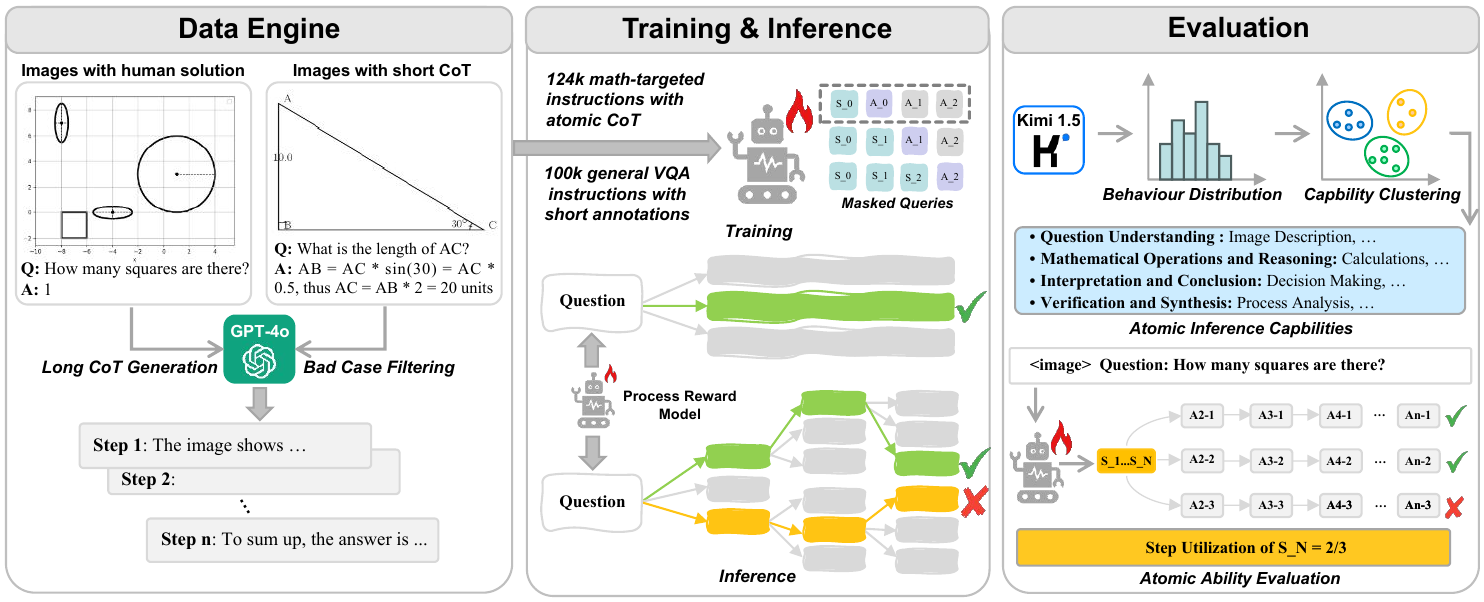} 
    \caption{The overview of AtomThink framework. We annotate and filter the open-source data with long CoT to generate atomic steps for fine-tuning and PRM training. During inference, step-wise or path-wise searching strategies can be applied to find optimal policies. Finally, the behavior distribution of GPT-4o is obtained through clustering with Kimi1.5, and an outcome-based method is employed for atomic step utilization evaluation.}
    \label{fig:framework}
\end{figure*}

\section{Method}

We present the details of AtomThink for promoting MLLMs in reasoning with self-structured CoT in this section. As shown in Figure \ref{fig:framework}, AtomThink consists of four key components, including a self-structured reasoning mechanism (Sec.~\ref{seq:scot}), a data engine (Sec. \ref{sec:engine}), a atomic step fine-tuning process (Sec. \ref{sec:atomic_finetune}) and an atomic capability evaluation (Sec. \ref{sec:metric}). 

\subsection{Self-structured Chain-of-Thought}
\label{seq:scot}
To enable MLLMs to adaptively generate diverse reasoning paths in response to varying problems, akin to human cognition, we have proposed an inference method based on Self-structured Chain-of-Thought (SCoT). In contrast to structured methodologies, our approach does not constrain the model to a fixed template of thought or a predetermined sequence of reasoning steps. Instead, it empowers the model to autonomously seek out the most appropriate reasoning behaviors during the inference process.

\paragraph{Multi-round Atomic Step Generation} We commence by defining the minimal predictive action with semantic consistency as an \textbf{Atomic Step}, which may constitute a single sentence or a combination thereof. Utilizing atomic steps as fundamental building blocks, we propose a multi-round prediction method to iteratively self-generate thought chains with dynamic structures. During the reasoning process, we prompt the model to predict only one minimal atomic step at a time, thereby focusing on the quality of each atomic step. Subsequently, the current prediction is appended to the historical reasoning steps and provided as contextual input for the next prediction cycle. Our reasoning template with SCoT is shown in the Appendix Figure~\ref{fig:prompts_atomthink}.

Due to the constraints of model size, we find that current MLLMs exhibit anomalies like reasoning stagnation and generating hallucinated content. Therefore, we use the following methods for anomaly detection and thought restart:
\begin{itemize}
  \item \textbf{Rule-based Filter.} We employ Jaccard similarity to quantify intra- and inter-step semantic repetition, thereby mitigating thought blockages and looping phenomena. Additionally, we define a \textit{max\_length} parameter to control the maximum length of atomic steps.
  \item \textbf{Temperature Accumulation.} Upon detection of an anomaly, we will perform a single-step inference anew to replace the erroneous atomic step. To enhance diversity of outcomes, we incrementally increase the temperature with each error to simulate the divergent thinking characteristic of human cognition.
\end{itemize}

\paragraph{Policy Search with Process Reward Model}
Given that the model spontaneously segments atomic steps during reasoning, a natural consideration is the introduction of a process reward model (PRM) to further expand the search space for predictive actions. As there are many search strategies to generate candidate actions, we categorize the existing strategies into path-wise searching and step-wise searching:

\begin{itemize}
  \item \textbf{Path-wise Search.} 1) Majority voting combines multiple reasoning paths by selecting the most frequent outcome across them. 2) Best-of-N generates \(C\) candidate rollouts simultaneously and selects the solution with the highest score. The score can be calculated by aggregating the overall value of the entire path.
  \item \textbf{Step-wise Search.} 1) Greedy algorithm focuses on making the locally optimal choice at each step of the reasoning process. 2) Beam search explores multiple branches at each step and maintains a fixed number of top candidates for each stage of reasoning. It balances between exploring different paths and exploiting the most promising ones.
\end{itemize}

In our principal experiments, we employed a step-wise beam search to extend the test-time. The Appendix~\ref{prm} and Table~\ref{tab:search} provides a detailed description and comparative experiments of different policy search methods.

\begin{table}[ht]
\centering
\resizebox{\columnwidth}{!}{  
\begin{tabular}{l c c c}
\toprule
\textbf{Source} & \textbf{AMATH-Metadata} & \textbf{AMATH-SFT} \\
\midrule
CLEVR        & 2056  & 11.9K   \\
Geometry3K   & 1224  & 9.3K  \\
MAVIS        & 1685  & 11.4K  \\
TabMWP       & 2643  & 16.3K  \\
GeomVerse    & 1347  & 9.9K \\
MathV360K    & 5632 & 31.6K  \\
GeoQA+       & 2222  & 15.5K  \\
IconQA       & 3199  & 18.1K \\
\midrule
Total        & 20008 & 124K \\
\bottomrule
\end{tabular}
}
\caption{Data composition of our AMATH. 20K VQA samples are applied to generate 124K SFT data with intermediate atomic steps.}
\label{tab:dataset}
\end{table}

\begin{table}[ht]
\centering
\begin{tabular}{c|c|c}
\hline
\textbf{Data} & \textbf{GPT Score} & \textbf{Avg. Length} \\
\hline
PRM800k & 84.1 & 1245.4 \\
Direct & 1.5 & 3.6 \\
Vanilla CoT & 79.6 & 670.5 \\
AMATH(Ours) & 89.4 & 849.8 \\
\hline
\end{tabular}
\caption{Comparison of different data styles. AMATH achieves the highest GPT-4o preference score and generates longer content than vanilla CoT.}
\label{tab:diff_data_style}
\end{table}

\subsection{Data Engine}
\label{sec:engine}
Guiding MLLMs toward deep reasoning requires a substantial amount of high-quality CoT data. However, in the field of visual mathematics, the scarcity of publicly available datasets presents a considerable challenge. To overcome this, we develop an automated data engine capable of generating step-by-step long CoTs, resulting in our own atomic multimodal dataset, dubbed AMATH. Specifically, our data engine introduces a dynamic prompting strategy and short CoT augmentation strategy to produce multi-step reasoning paths. Subsequently, we propose a difficulty scoring mechanism coupled with a secondary review strategy to sift through and filter out erroneous instances.


\paragraph{Multimodal CoT Generation.} For long CoT generation, we propose two prompt-based methods:
\begin{itemize}
  \item \textbf{Dynamic Prompting.} Inspired by recent research \cite{g1}, we propose a dynamic prompt strategy for generating atomic inference steps. Specifically, our strategy drives a LLM to iteratively construct state-reasoning paths. Each path node represents a reasoning step and encompasses the previous stage, the current state, and a possible action. The possible action includes continuing reasoning, verifying, and drawing conclusion, which is determined by the LLM itself. The prompt is shown in the Appendix.
  \item \textbf{Short CoT Augmentation.} To fully leverage existing short CoT annotations of VQA datasets, we also employ an MLLM to atomize and augment these annotations. This approach allows us to semantically segment an original reasoning process into multiple discrete steps, and focus on solving a single atomic problem at each stage of the reasoning process.
\end{itemize}

\paragraph{Bad Case Filtering.} Due to the prevalence of substantial noise within the publicly available datasets, we first employ a difficulty scoring system to filter the questions. Subsequently, a LLM is used for a secondary review to eliminate erroneous CoTs.
\begin{itemize}
  \item \textbf{Difficulty Scoring.} To quantify the difficulty of questions, we employ Qwen2-VL-7B to sample $N$ candidates for each question, using the win rate of $N$ candidates as difficulty level of the question ($N=10$ in our paper). To enhance the efficiency of training, we have removed most questions with a difficulty level of 0.
  \item \textbf{Secondary Review.} Upon the generation of CoT, we utilize GPT-4o to conduct secondary review, with a particular focus on the accuracy of atomic steps and the correctness of final answers. Furthermore, we engage two professional annotators to perform a sampling inspection of our dataset.

\end{itemize}

\begin{figure*}[t]
    \centering
    \includegraphics[width=\textwidth]{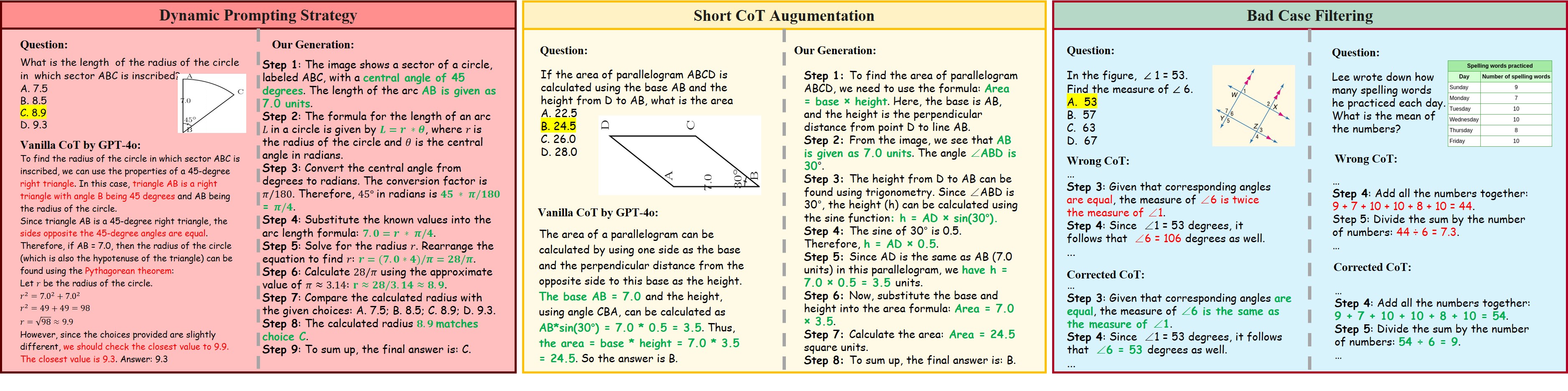} 
    \caption{A case study of our data engine to generate high quality CoT. Red and green characters denote incorrect and correct responses, respectively. Compared with vanilla CoT generated by GPT-4o, our dynamic prompting strategy exhibits fewer hallucinations in every atomic steps. Utilizing existing short annotations, we can augment longer paths that encompass more details. Additionally, bad case filtering is applied to inspect low-quality noisy data within the automated pipeline.}
    \label{fig:case}
\end{figure*}

\paragraph{AMATH Dataset.} 
We sample multimodal reasoning data from CLEVR \cite{johnson2017clevr}, Geometry3K \cite{geo3k}, MAVIS \cite{zhang2024mavis}, TabMWP \cite{TabMWP}, GeomVerse \cite{kazemi2024geomverse}, Mathv360k \cite{shi2024math}, GeoQA+~\cite{chen2021geoqa} and IconQA~\cite{lu2021iconqa}. For GeomVerse and MAVIS, we conduct short CoT augmentation, while the rest are generated by dynamic prompts to produce multi-step reasoning. Table \ref{tab:dataset} illustrates the distribution of our data. In Table \ref{tab:diff_data_style}, we also evaluate the quality in a subset of 500 AMATH samples with GPT-4o scoring. We have illustrated the generation and filtration examples of our dataset in Fig.~\ref{fig:case}.

\subsection{Atomic Step Fine-Tuning}
\label{sec:atomic_finetune}

To fully exploit MLLMs for addressing multi-modal mathematical problems, we conduct fine-tuning with atomic step-wise reasoning. We have dissected CoTs from the metadata of AMATH into atomic steps and subsequently employed serialized masking to incrementally incorporate these into the historical reasoning steps, thereby generating multiple training samples (denoted as AMATH-SFT) for supervised instruction fine-tuning.

\subsection{Atomic Capability Evaluation}
\label{sec:metric}

Similar to human problem-solving processes, a SCoT may involve multiple reasoning abilities. However, traditional CoT methods do not focus on the ability to follow individual reasoning step or provide fine-grained analyses of the underlying abilities. To address this gap, we have developed an atomic capability evaluation strategy, offering a new analytical perspective for reasoning.

Our evaluation method aims to assess the mathematical capabilities of a target model from various perspectives, such as understanding, operations, and certifications. To this end, we first construct a canonical set of capabilities. As shown in Figure \ref{fig:pie_outline}, we collect the behavior distribution of GPT-4o on AMATH dataset and use Kimi-1.5 to perform clustering, yielding clusters that each of them represents a certain ability utilized by high-level intelligent models in solving mathematical problems. We consider each cluster as a set and let $Set(a)$ denote the cluster of an ability $a$.

We initially posit that models with superior atomic reasoning capabilities are more adept at leveraging recent contextual steps to further excavate answers. Hence, we can quantify a certain reasoning ability of a model based on its average probability of reaching a correct answer with its rollouts sampled from the corresponding ability set. Specifically, assume a question has $n$ historical reasoning steps $S=\{s_i|i =1,...,n)\}$. We define the step utilization rate $u(S)$ as the probability of reaching an answer by continuing to reason based on $S$ averaging on $M$ sampled rollouts:
 \begin{align}
u(S) = \frac{\sum^M_{m=1}\llbracket {r}_{m} \text{ is correct}\rrbracket}{M},
 \label{rate}
\end{align}
where $r_m$ is the $m$-th rollout. Subsequently, we calculate the utilization rates of different historical steps and map the corresponding $S$ back to the set of atomic capabilities. We compute the average utilization rate for each category in the ability set to represent the model's atomic reasoning capability, which can be represented as follows,
 \begin{align}
 Score(a) = \frac{1}{{|Set(a)|}}\sum\limits_{{S_k} \in Set(a)} {u({S_k})} .
 \label{ability_score}
\end{align}
In our experiments, we selected 160 samples from an out-of-distribution mathematical dataset (R1V-Stratos \cite{yu25r1vision}), to construct a test set for atomic capability evaluation.


\section{Experiment}
\label{sec:exp}

\begin{table*}[h!]
\centering
\begin{tabular}{
    c 
    c 
    |
    c 
    c 
    |
    c 
    |
    c 
    |
    c
}
\toprule
\textbf{Model} & \textbf{Inference} &\textbf{MathVista-M} & \textbf{MathVista} & \textbf{MathVerse} & \textbf{MathVision} & \textbf{HLE} \\

\midrule
Random Choice & - & - & 17.9 & 12.4 & 7.2 & -\\
Human & - & - & 70.9 & - & 68.8& - \\
OpenAI o1 & CoT & - &73.9 & -&- & 8.8\\
Claude 3.5 Sonnet &CoT&-&67.7&-&38.0&4.8\\
GPT-4o & CoT & - &63.8 & -&- &3.1\\
GPT-4V & CoT & - &49.9& 54.4 & 24.0 & -\\
LLaVA-NeXT-34B & Direct & -& 46.5  & 23.8  & -& -\\
InternLM-XComposer2 & Direct&- & 57.6 & 16.5 & 14.5& -\\
Qwen-VL-Plus & Direct & -&43.3  &11.8 & 10.7 & -\\
LLaVA-1.5-13B &Direct&-&27.6 & 15.6&11.2& -\\
G-LLaVA-7B &Direct&-&53.4&16.6&- & -\\
MAVIS-7B & Direct & -&29.1&27.5&19.2 & -\\
LLamaV-o1-11B & CoT & -&54.4&-&- & -\\
LLaVA-CoT-11B & CoT & -&54.8&-&- & -\\
\midrule
LLaVA1.5-7B*& Direct &23.3& 27.9&10.0&9.3& 4.2\\
AtomThink-LLaVA & SCoT & 
  26.5{\textcolor{improvegreen}{\small\ (+3.2)}} & 
  29.2{\textcolor{improvegreen}{\small\ (+1.3)}} & 
  14.4{\textcolor{improvegreen}{\small\ (+4.4)}} & 
 12.7{ \textcolor{improvegreen}{\small\ (+3.4)}} &
 5.7{ \textcolor{improvegreen}{\small\ (+1.5)}} \\
AtomThink-LLaVA & SCoT w/ PRM & 
  31.1\textcolor{improvegreen}{{\small\ (+7.8)}} & 
  32.1\textcolor{improvegreen}{{\small\ (+4.2)}} & 
  14.6\textcolor{improvegreen}{{\small\ (+4.6)}} & 
  12.3\textcolor{improvegreen}{{\small\ (+3.0)}} & 
  4.4\textcolor{improvegreen}{{\small\ (+0.2)}} \\
\midrule
Llama3.2-Vision-11B*& Direct &44.1&47.5&23.3&13.8& 4.0 \\
AtomThink-LlamaV & SCoT & 
  56.9\textcolor{improvegreen}{{\small\ (+12.8)}} & 
  57.1\textcolor{improvegreen}{{\small\ (+9.6)}} & 
  31.5\textcolor{improvegreen}{{\small\ (+8.2)}} & 
  18.2\textcolor{improvegreen}{{\small\ (+4.4)}} &
  5.4\textcolor{improvegreen}{{\small\ (+1.4)}} \\
AtomThink-LlamaV & SCoT w/ PRM & 
  59.1\textcolor{improvegreen}{{\small\ (+15.0)}} & 
  58.4\textcolor{improvegreen}{{\small\ (+10.9)}} & 
  33.5\textcolor{improvegreen}{{\small\ (+10.2)}} & 
 21.0\textcolor{improvegreen}{{\small\ (+7.2)}} &
 4.5\textcolor{improvegreen}{{\small\ (+0.5)}} \\

\bottomrule
\end{tabular}
\caption{Comparison of accuracy with state-of-the-art models on four benchmarks. Our AtomThink achieves consistent improvement across models of varying scales and surpasses baselines on all four benchmarks. Specially, AtomThink-LlamaV, with 11B parameters, surpasses GPT-4V by 8.5\% on MathVista. The baseline models (*) are post-trained by LLaVA100K VQA.}
\label{tab:main}
\end{table*}

\begin{figure}[ht]
    \centering
    \includegraphics[width=0.5\textwidth]{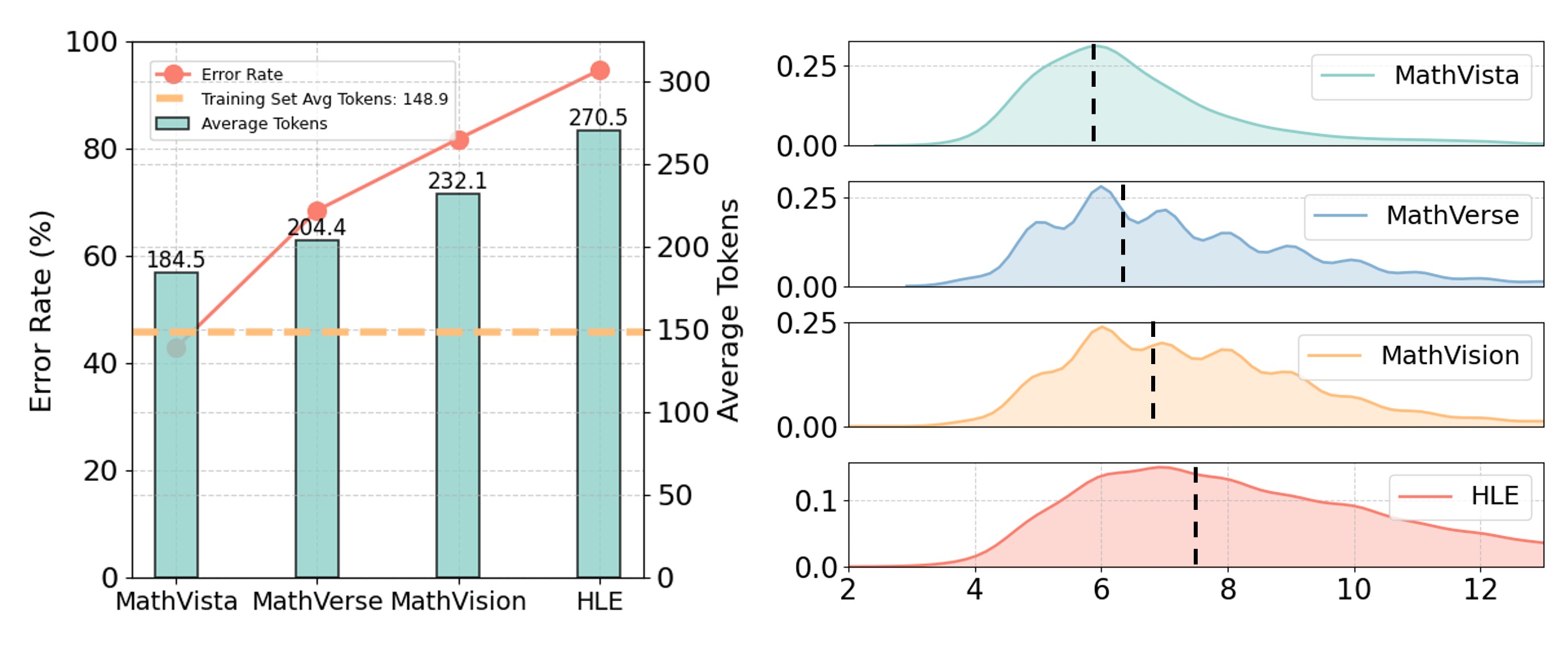} 
    \caption{Comparison of the average response length in AtomThink-LlamaV over benchmarks with different complexity. (a) As tasks become more challenging, the model proactively utilizes more tokens. (b) The proportion of longer CoT containing a greater number of atomic steps increases in outputs.}
    \label{fig:step_length}
\end{figure}

\begin{figure}[ht]
    \centering
    \includegraphics[width=0.5\textwidth]{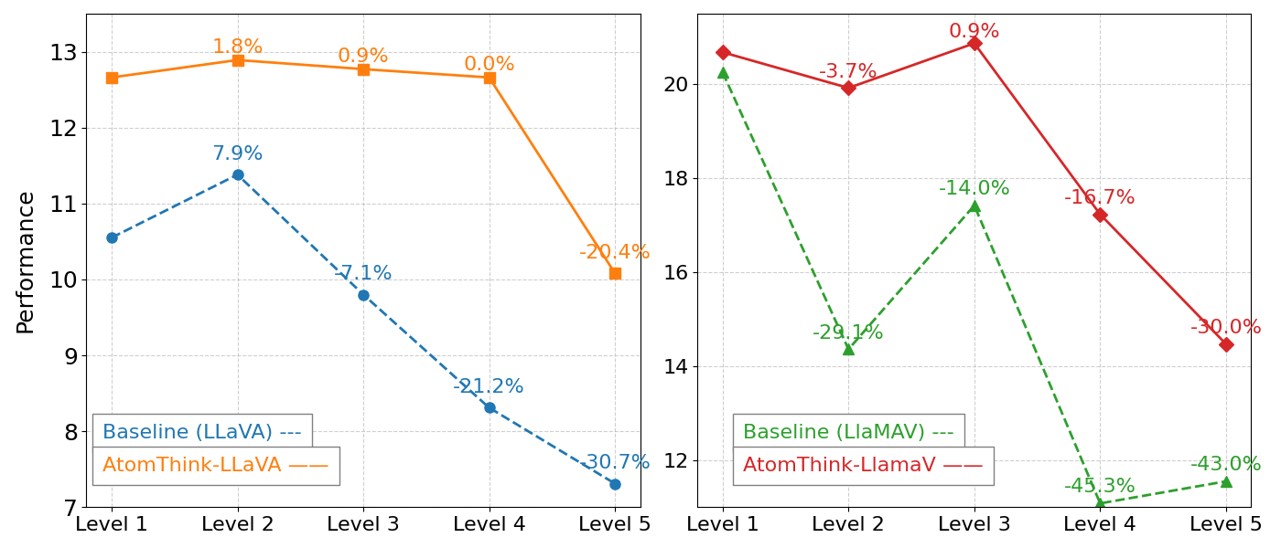} 
    \caption{MathVision-mini accuracy in diverse difficulty level subsets. A higher level signifies increased difficulty. The performance decline margin of AtomThink modes are more narrow (-20.4\% v.s. -30.7\% in LLaVA1.5, -30\% v.s. -43.0\% in LlamaV).}
    \label{fig:difficulty_performance}
\end{figure}

\begin{table*}[ht]
\centering
\begin{minipage}{0.75\textwidth} 
\centering
\resizebox{\columnwidth}{!}{  
\begin{tabular}{l c c c}
\toprule
\textbf{Method} & \textbf{LLaVA-CoT}  & \textbf{AtomThink-LlamaV} & \textbf{AtomThink w/ PRM} \\
\midrule
Accuracy        & 54.8 & 57.1\textcolor{improvegreen}{{\small\ (+2.3)}}& 58.4\textcolor{improvegreen}{{\small\ (+3.6)}}\\
Dataset Scale   &  100k & 20k\textcolor{improvegreen}{{\small\ (-80\%)}}& 20k\textcolor{improvegreen}{{\small\ (-80\%)}} \\
Tokens     & 1322.2 & 161.5\textcolor{improvegreen}{{\small\ (-87.8\%)}} & 734.7\textcolor{improvegreen}{{\small\ (-44.4\%)}} \\
Inference Time     & 57.2 & 8.4\textcolor{improvegreen}{{\small\ (-85.3\%)}} & 38.1\textcolor{improvegreen}{{\small\ (-33.4\%)}} \\
\bottomrule
\end{tabular}
}
\caption{Comparison with LLaVA-CoT. We not only improve inference accuracy by 3.6\%, but also decrease the data and test-time resource requirement.}
\label{tab:llavacot}
\end{minipage}
\hfill
\begin{minipage}{0.2\textwidth} 
\centering
\resizebox{\columnwidth}{!}{  
\begin{tabular}{l c}
\toprule
\textbf{Samples} & \textbf{Accuracy} \\
\midrule
0        & 9.28\\
10k   &   9.67 \\
30k    &   9.33 \\
60k       &   11.33 \\
90k    &  8.97 \\
124k    &  12.45 \\
\bottomrule
\end{tabular}
}
\caption{AtomThink-LLaVA performance improvement of MathVision-mini with dataset scaling.}
\label{tab:data_scaling}
\end{minipage}
\end{table*}

\begin{table}[ht]
\centering
\resizebox{0.8\columnwidth}{!}{  
\begin{tabular}{c c c}
\toprule
\textbf{Candidate} & \textbf{Ouput Tokens} & \textbf{Accuracy}\\
\midrule
0        & 2.3  & 13.9  \\
1   & 231.9  & 18 \\
2        & 518.6  & 18.3 \\
3       & 822.3  &23.3 \\
\bottomrule
\end{tabular}
}
\caption{AtomThink-LlamaV performance improvement of MathVision-mini with test-time scaling. We employ Best-of-N and PRM to select the optimal step among N candidates.}
\label{tab:tt_scaling}
\end{table}

\subsection{Setup}
\paragraph{Baselines.} Our experiments utilize two open-source MLLMs, including LLaVA1.5-7B~\cite{liu2024visual} and Llama3.2-11B-Vision~\cite{llama32v}. With a subset of 100K multimodal question-answer pairs sampled from LLaVA-665K~\cite{liu2024visual}, we post-training full parameters of their language models, projectors and vision encoder as baselines. For our AtomThink models, the AMATH-SFT dataset introduced in Section \ref{sec:engine}, is incorporated to introduce atomic reasoning capabilities. We use a learning rate of 2e-6 and a batch size of 128 to fine-tune for one epoch. We select 12 cutting-edge MLLMs for comparison, including Claude 3.5 Sonnet~\cite{claude}, OpenAI’s o1~\cite{openai2024_o1}, 4o~\cite{openai2024_4o}, 4v~\cite{openai2024_4v}, as well as LLava-NeXT-34B~\cite{liu2024llavanext}, InternLM-XComposer2~\cite{zhang2024mavis}, Qwen-VL-Plus~\cite{bai2023qwen}, LLaVA-1.5-13B~\cite{liu2024visual}, GLLaVA-7B~\cite{gao2023gllava}, MAVIS-7B~\cite{zhang2024mavis}, LlamaV-o1-11B~\cite{llamavo1} and LLaVA-CoT-11B~\cite{llavacot}.

\paragraph{Evaluation Setting.}
We evaluated the performance of our method on MathVista~\cite{lu2023mathvista}, a publicly available benchmark encompassing both general-targeted and mathematics-targeted (MathVista-M) domains. Additionally, MathVerse~\cite{zhang2025mathverse} is introduced to assess the model's sensitivity to mathematical graphs. MathVision~\cite{wang2024measuring}, a benchmark encompassing a diverse range of mathematical problem complexities, is also incorporated into the experiments to specifically evaluate the dynamic variations in our atomic steps. We also introduced Humanity's Last Exam (HLE)~\cite{phan2025humanity}, one of the most challenging benchmark, to assess the model's reasoning capabilities under extremely difficult conditions.

Our evaluations include four inference settings, including \textbf{Direct}, \textbf{CoT}, \textbf{SCoT}, and \textbf{SCoT w/ PRM}. In the \textbf{Direct} setting, we prompt the model to generate a concise final answer. In \textbf{CoT}, the model is instructed to answer the question through step-by-step reasoning. For the Direct and CoT evaluations, we use prompts from lmms-eval~\cite{zhang2024lmmsevalrealitycheckevaluation,lmms_eval2024}. Our AtomThink-models support two additional settings: \textbf{SCoT} and \textbf{SCoT w/ PRM}. In SCoT, our models follow a single, atomic reasoning path based purely on their learned policies, without employing any supplementary search strategies. In SCoT w/ PRM, enhanced by Qwen2.5-Math-PRM-7B~\cite{prmlessons}, we utilize step-wise beam search with a window of 3 and candidate number of 2. During the search process, the temperature for each step is initialized at 0 and incremented by 0.5 with each candidate sampling to enhance diversity.


\begin{figure}[ht]
    \centering
    \includegraphics[width=0.5\textwidth]{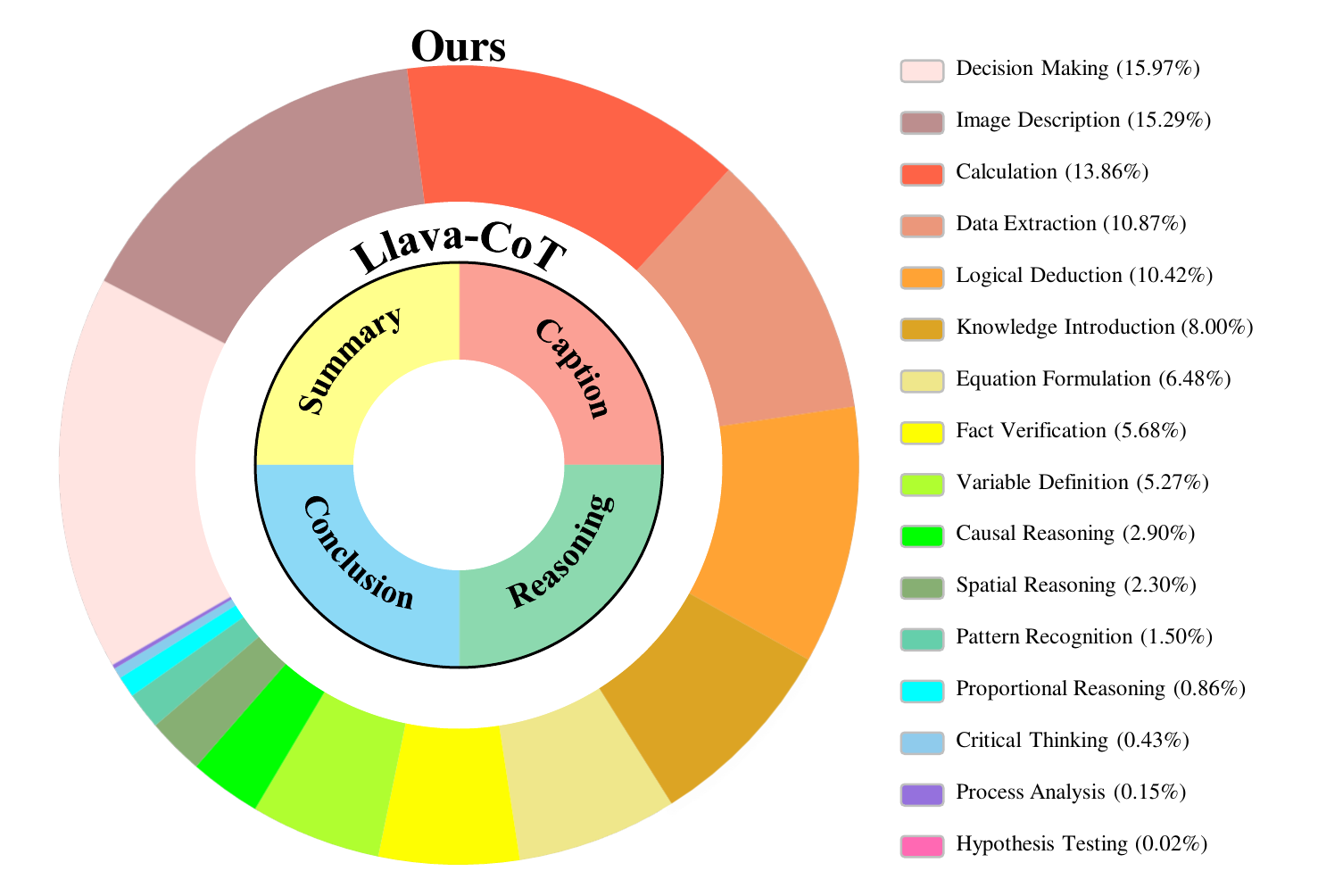} 
    \caption{Atomic step distribution of AtomThink-LlamaV in R1V-Stratos-160 testset. Compared with the structured method, our model is capable of performing a variety of reasoning behaviors.}
    \label{fig:pie_outline}
\end{figure}

\begin{figure}[ht]
    \centering
    \includegraphics[width=0.5\textwidth]{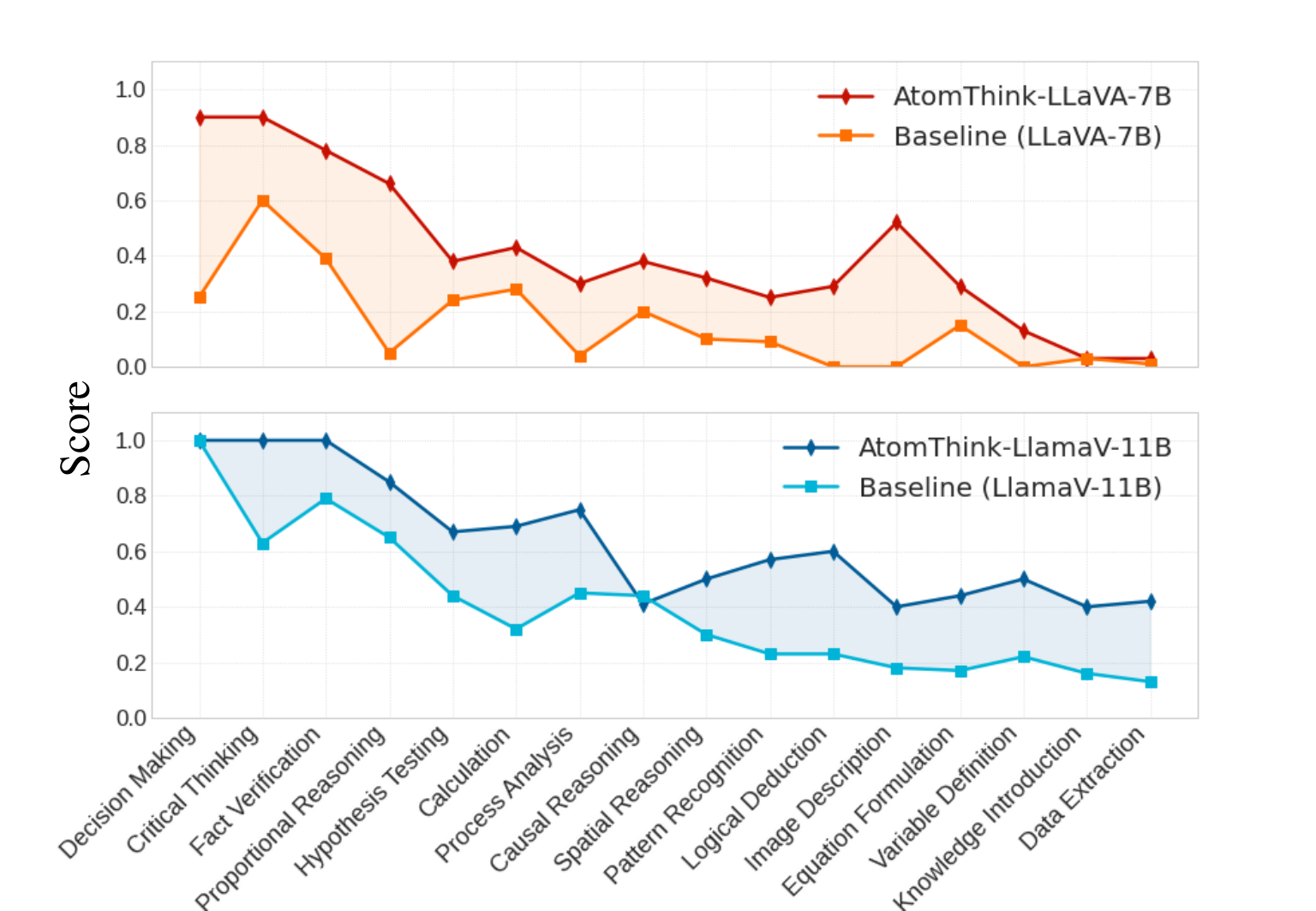} 
    \caption{The evaluation of utilization efficiency across different atomic capabilities. Lower scores in initial stages of CoT (e.g. Data Extraction) indicate the presence of error accumulation.}
    \label{fig:score_outline}
\end{figure}

\subsection{Main Results}

In Table~\ref{tab:main}, our AtomThink framework is applied to train LLaVA1.5-8B and Llama3.2-Vision-11B, yielding consistent performance improvements over the original models. With Self-structured CoT, the accuracy of AtomThink-LLaVA can be enhanced by 4.4\% and 3.4\% in MathVerse and MathVision, respectively. In a larger vision understanding model, AtomThink-LlamaV gains a higher improvement by 9.6\% and 8.2\%. When combined with step-wise beam search and process reward model, AtomThink-LlamaV achieves a new state-of-the-art on MathVista, surpassing GPT-4V and narrowing the gap between MLLMs and human performance. Furthermore, in the more challenging and complex HLE task, despite being unable to answer the majority of questions, our two models still achieved modest improvements (1.5\% and 1.4\%). Overall, we increase the average performance of 7B and 11B models by 4\% and 8.8\%, respectively. These results demonstrate the framework's strong generalization capability and practical usability.

\subsection{Scaling Reasoning According to Difficulty}
To assess the variation in the length of unstructured CoT, we present the output distribution of AtomThink-LlamaV across four benchmarks in Fig.~\ref{fig:step_length}. The ascending error rates indicate a sequential increase in benchmark difficulty. In subplot (a), without human intervention, the model employs a greater number of atomic steps to address more complex problems. Surprisingly, despite the average token count of AMATH-SFT being only 148.9, the outputs of our model on different test sets are progressively increasing. This suggests that the model is not merely fitting the training data but is instead exhibiting an emergent ability to autonomously explore the depth of reasoning, potentially representing the \textbf{``Aha Moment"} we anticipate in visual reasoning models. Moreover, although accuracy decreases with increasing difficulty level, Fig.~\ref{fig:difficulty_performance} demonstrates that the decline margin is reduced when AtomThink is applied.

\subsection{Autonomous Generation of Diverse Structures}
 We cluster the reasoning behaviors of GPT-4o into 16 categories and collect the distribution of atomic steps produced by AtomThink on the Stratos160 test set. The results in Fig.~\ref{fig:pie_outline} demonstrate that, compared to structured output (LLaVA-CoT), our SCoT exhibits a more diverse range of reasoning structures. Among all categories, the higher proportion of Image Description (15.29\%) and Data Extraction (10.87\%) underscores the importance of perceptual capabilities. With the enhanced visual understanding abilities, the model also displays specific behaviors such as Causal Reasoning (2.9\%) and Spatial Reasoning (2.3\%).

\subsection{Data Utilization and Reasoning Efficiency}
In Table~\ref{tab:llavacot}, we present a comprehensive comparison with state-of-the-art methods in terms of accuracy, dataset scale, output token count, and inference time. By utilizing only one-fifth of VQA samples, we achieve a 3.6\% improvement on MathVista. Furthermore, due to our ability to provide concise responses to simpler questions, we reduce inference time by 85.3\% and 33.4\% per sample when not using search and employing PRM for strategic search, respectively.

\subsection{Scaling Law in Data and Test-time}
Previous research has found that scaling up data and test-time computations can enhance the reasoning in language models. Our result also discovers that this scaling law persists in multimodal models. Fig.~\ref{tab:data_scaling} shows that increasing data scale generally promotes performance. By employing a step-wise Best-of-N strategy, we linearly increased reasoning time, with each additional candidate improving accuracy by an average of 3.1\%.

\subsection{Further Analysis}
\paragraph{What Kind of Capabilities Do MLLM Need in Reasoning?}
Building upon the set of atomic capabilities illustrated in Fig.~\ref{fig:pie_outline}, we calculated our model's utilization rate for each category of steps using Eq. (~\ref{ability_score}). Results in Fig.~\ref{fig:score_outline} reveal that as the given historical steps approach the beginning of the reasoning chain (e.g. Image Description and Data Extraction), prediction error rate continuously increases. This error accumulation effect prompts us to focus on the quality of reasoning in initial stages. In future work, we can mitigate the rate of error accumulation by adjusting data ratios and designing sampling strategies.

\paragraph{What Kind of Information Do PRM Focus on?}
In Table~\ref{tab:main}, we find that even the reasoning process heavily relies on visual dominant inputs, the highest performance is achieved by using a language PRM. Meanwhile, we have tried to train a vision-language PRM of text and visual modalities data. Unfortunately, the results indicate that the language reward model achieves better performance than the MLLM. This suggests that the current PRM paradigm may not be adept at utilizing visual information. Exploring how to leverage multimodal features to correct the reasoning process will be a direction we need to investigate.

\section{Conclusion}
To mitigate overthinking and structured output, we have proposed a self-structured chain of thought method. It ensures reasoning efficiency while adaptively generating a variety of atomic steps. Subsequently, we introduced AtomThink, a comprehensive deep reasoning framework that encompasses data engineering, model training, inference, and evaluation. The experimental results demonstrate that our method consistently enhances the model's diverse behaviors during the problem-solving process, leading to improved reasoning performance across various multimodal benchmarks. This work paves the way for developing generalized slow-thinking models and provides novel insights for understanding multi-modal reasoning patterns.

{
    \small
    \bibliographystyle{ieeenat_fullname}
    \bibliography{main}
}
\clearpage
\setcounter{page}{1}
\maketitlesupplementary

\appendix  
\section{Implementation Details}
\subsection{Policy Models} 
In this section, we provide more implementation details for baseline models and our framework. we post-train them using AMATH-SFT and a sub-sampled dataset of LLaVA665K, containing 100k samples. During this process, the weights of LLM, projector and vision encoder are fully fine-tuned. Specifically, we utilize the Llama-factory framework to train the models and the hyperparameters are listed in Table\ref{tab:parameters}.

\begin{table}[ht]
    \centering
    \begin{tabular}{l|c|c}
    \toprule
    \textbf{Parameter} & \textbf{LLaVA1.5-7B} & \textbf{Llama3.2-V-11B} \\ 
    \midrule
    Learning Rate & 2e-6 & 2e-6 \\ 
    Epochs & 1 & 1 \\ 
    Batch Size & 128 & 128 \\ 
    Context Length & 4096 & 4096 \\ 
    Seed & 42 & 42 \\ 
    Precision & FP16 & BF16 \\ 
    GPU & 32 NVIDIA V100& 8 NVIDIA A800\\ 
    FSDP & True & True \\ 
    DeepSpeed & Zero3 & Zero3 \\
    \bottomrule
    \end{tabular}
    \caption{Comparison of Parameters for post-training LLaVA1.5-7B and Llama-3.2-Vision-11B.}
    \label{tab:parameters}
\end{table}

\subsection{Atomic Search with PRM.} 
\label{prm}
With the fine-tuned MLLM capable of atomic step reasoning, we apply well-trained PRM, Qwen2.5-Math-PRM-7B~\cite{prmlessons}, for providing feedback. As there are many search strategies to generate candidate actions, we categorize the existing strategies into path-wise searching and step-wise searching and explore them in our AtomThink framework. Unlike traditional token-based search strategies, we sample candidates using atomic steps as the fundamental unit. 

\paragraph{Path-wise Search.} In path-wise searching, we build upon prior work \cite{wang2024openr,snell2024scaling} by parallel sampling multiple paths and aggregating scores to find optimal solutions. We investigate the following two strategies:
\begin{itemize}
  \item \textbf{Majority Voting: } It combines multiple reasoning paths by selecting the most frequent outcome across them. It assumes that the consensus across different paths is more likely to lead to the correct answer.
  \item \textbf{Best-of-N:} Given a generative MLLM, the best-of-N sampling method generates \(C\) candidate rollouts simultaneously and selects the solution with the highest score. The evaluation of candidate reasoning processes is determined by the PRM, which employs three aggregation methods to map the dense scores to the overall value of the entire path: \textbf{1) The worst action: } Compare the worst action among all candidate rollouts. It penalizes solutions with any weak action and is used to search a reasoning that is sensitive to errors. \textbf{2) The last action:} The score is derived from the prediction of the final answer in inference. \textbf{3) Average score:} It is calculated by averaging the rewards of all the actions in a chain. The explainability and consistency of intermediate reasoning are emphasized here as important as the outcome.

\end{itemize}
\paragraph{Step-wise Search.} Searching strategies of this type start with an initial path and incrementally expand the sampling space for each atomic action. Beam search and greedy strategies are applied to prune branches with low quality.
\begin{itemize}
  \item \textbf{Greedy Algorithm:} It focuses on making the locally optimal choice at each step of the reasoning process. It selects the best immediate action (step) based on the current state, without considering future consequences. 
  \item \textbf{Beam Search:} It explores multiple branches at each action and maintains a fixed number of top candidates for each stage of reasoning. It balances between exploring different paths and exploiting the most promising ones.
\end{itemize}

\section{Attempts on training a R1-like MLLM}
\begin{figure}[ht]
    \centering
    \includegraphics[width=0.5\textwidth]{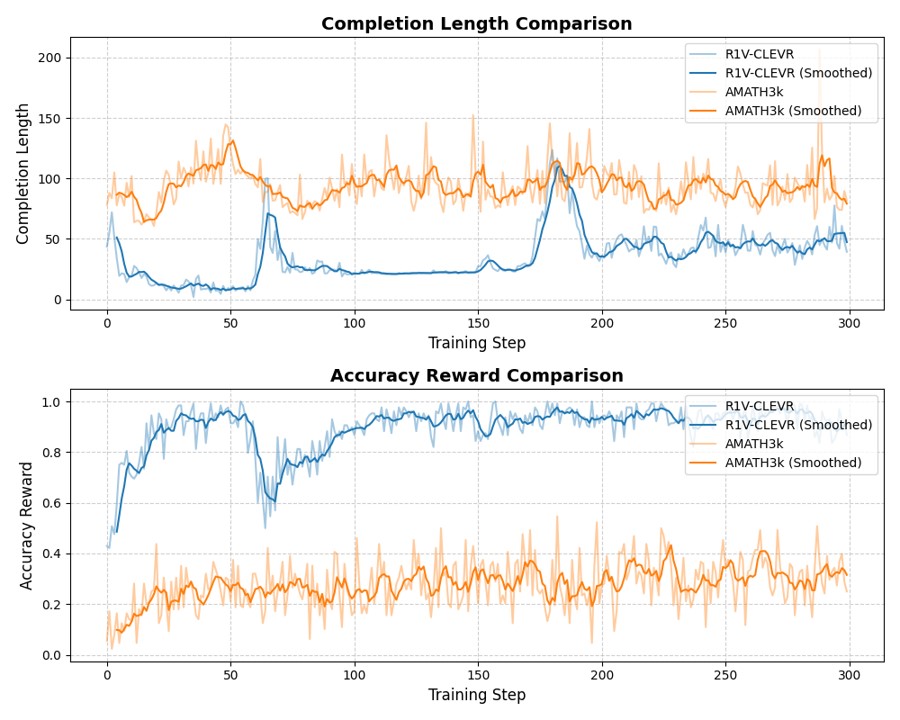} 
    \caption{Comparison with a DeepSeek-R1 like framework using reinforcement learning. A 3k subset of AMATH is sampled for fair comparison.}
    \label{fig:grpo}
\end{figure}

Recently, the introduction of DeepSeek-R1~\cite{guo2025deepseek} has demonstrated that reinforcement learning can effectively enhance autonomous reasoning capabilities. GRPO (Generalized Reward Prediction Optimization) as it's main training strategy, focuses on optimizing the prediction of rewards in complex environments to improve decision-making and policy learning. Several outstanding open-source repositories have explored the application of GRPO in the visual domain~\cite{yu25r1vision,openr1m}.

With the setup of R1V~\cite{yu25r1vision}, we also attempt to use GRPO to address complex reasoning problems. To facilitate a fair comparison, we sample a 3K subset from AMATH-metadata, equivalent in scale to R1V-CLEVR, and conducted experiments on Qwen2-VL-2B. The reward function is divided into format rewards and accuracy rewards. The Figure~\ref{fig:grpo} illustrates the changes in candidate length and accuracy rewards during the training process. Due to the use of more diverse and complex training data, we generate longer CoT completion length and lead to higher error rates. Although there is an initial improvement in accuracy rewards during the early stages of AMATH3K training, it plateaues around 30\%. Additionally, neither set of experiments exhibit the CoT growth phenomenon observed in R1. We think that pure reinforcement learning methods face greater challenges in multimodal tasks, potentially relate to task difficulty, model foundational capabilities and reward settings.

\section{Prompts Design}
\begin{figure*}[ht]
    \centering
    \includegraphics[width=0.9\textwidth]{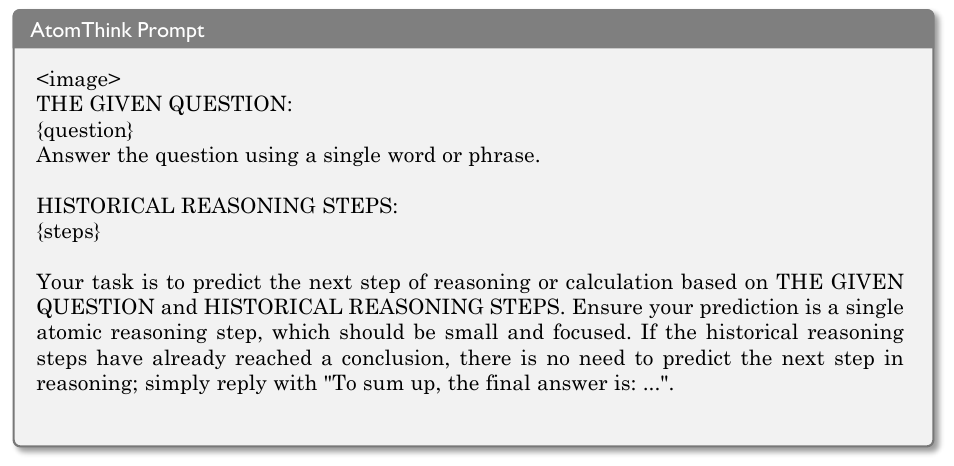} 
    \caption{AtomThink template for generating Self-structured CoT. The model takes an image and a question as input, generating an atomic step at each iteration. These steps are then concatenated into the historical reasoning steps, which are fed into model for the next round of reasoning.}
    \label{fig:prompts_atomthink}
\end{figure*}

\begin{figure*}[t]
    \centering
    \includegraphics[width=0.9\textwidth]{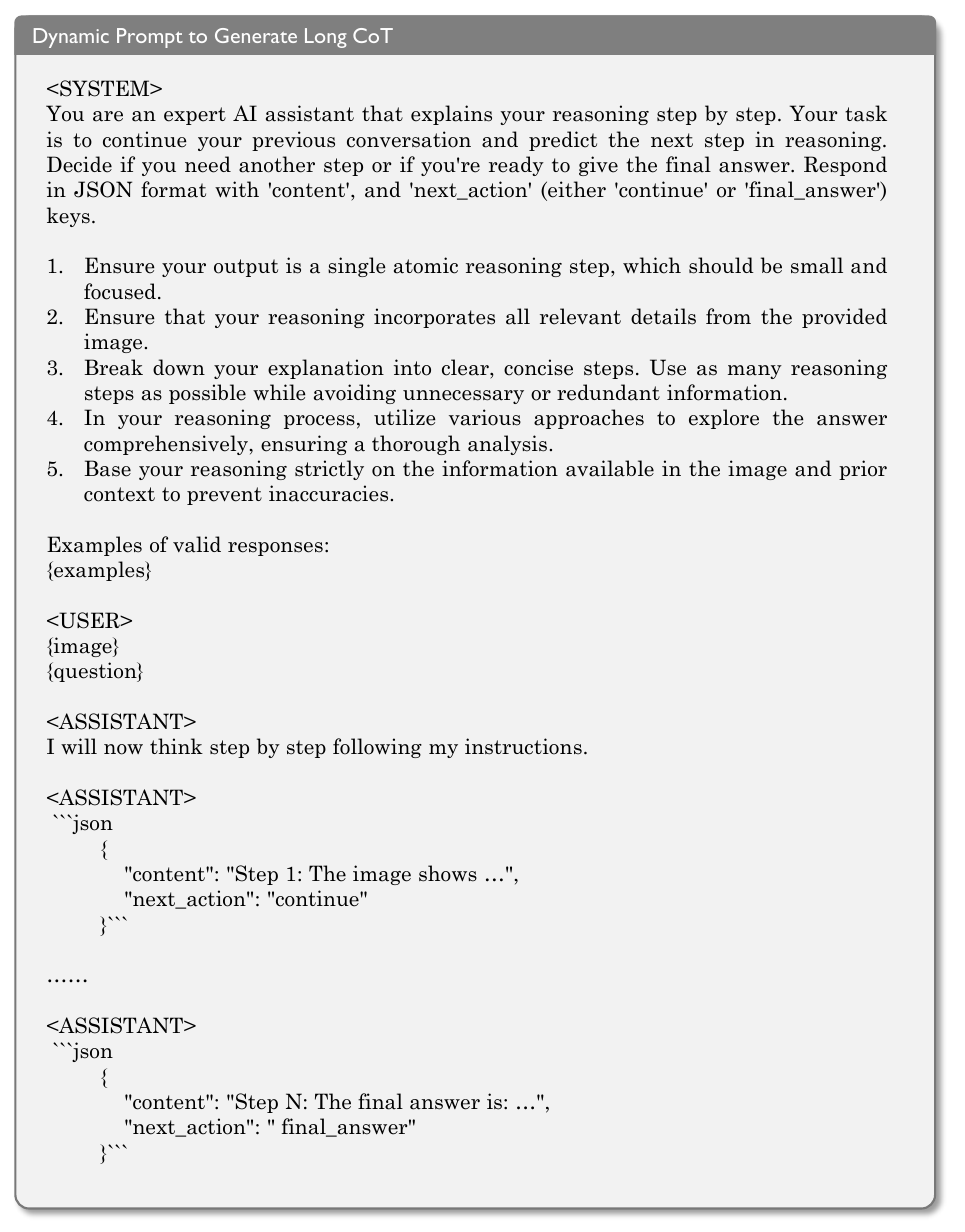} 
    \caption{Dynamic prompt for long CoT generation. Inspired by previous work, we designed a dynamic prompt template that generates reasoning steps for each iteration. It effectively identifies the input visual information to generate detailed image captions and fine-grained atomic steps.}
    \label{fig:prompts_dynamic}
\end{figure*}

\begin{figure*}[t]
    \centering
    \includegraphics[width=\textwidth]{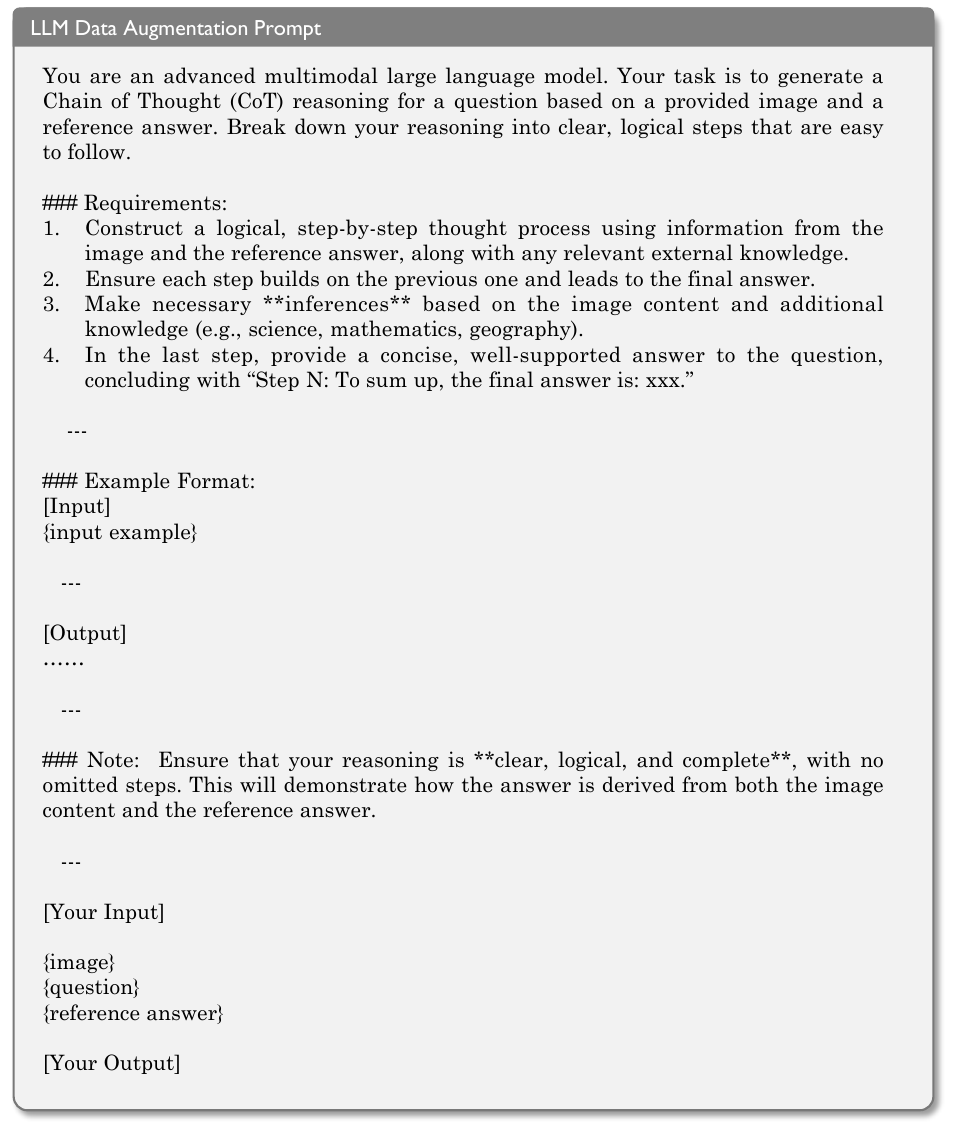} 
    \caption{Prompt for short answer augmentation. Using the current math VQA dataset, which already includes short answers and CoTs, we apply this template to enhance and generate detail atomic steps.}
    \label{fig:prompts_cot_aug}
\end{figure*}

\begin{figure*}[t]
    \centering
    \includegraphics[width=\textwidth]{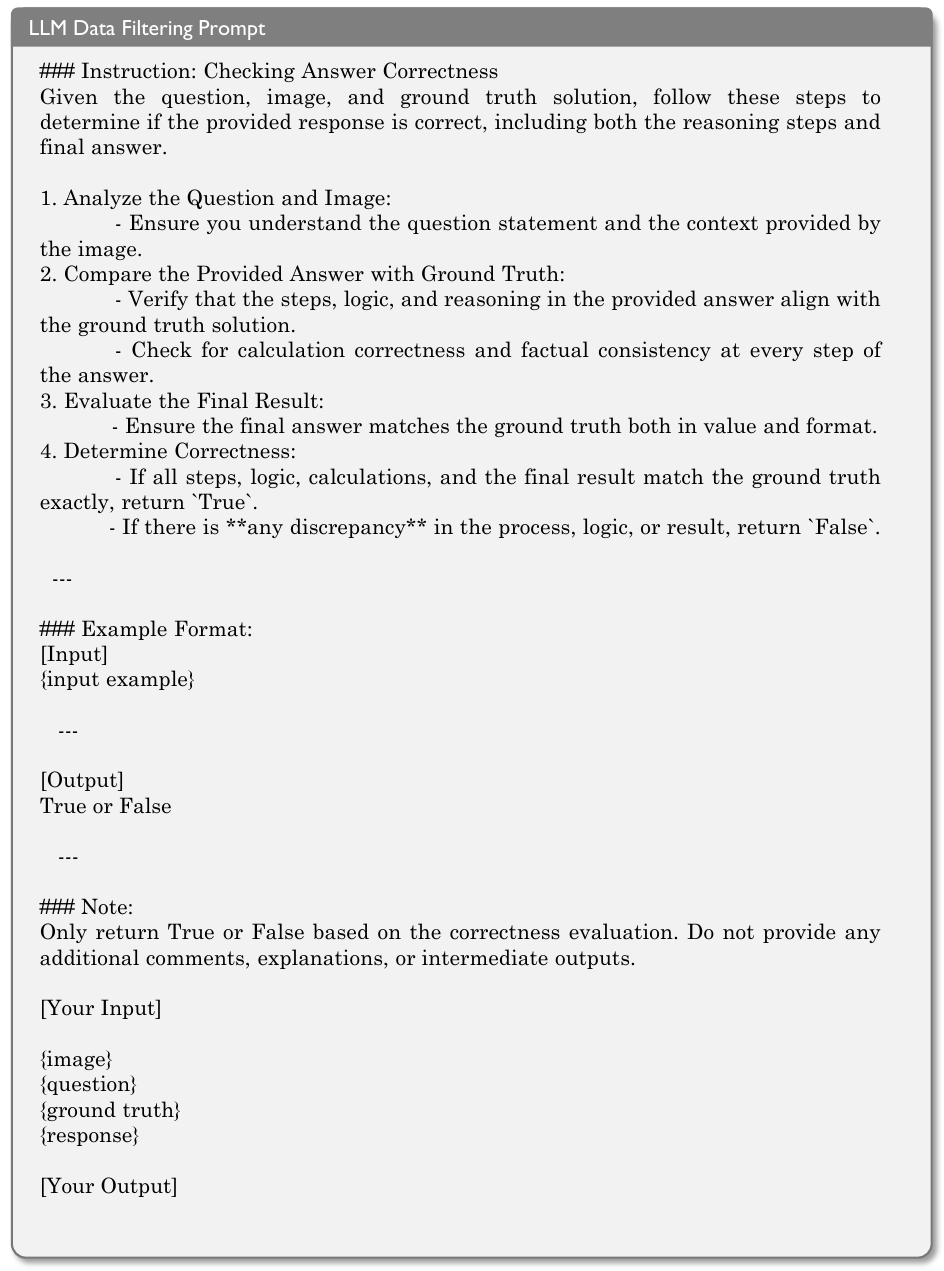} 
    \caption{Prompt for filtering wrong CoT. Due to the quality gap between the reasoning steps generated by the AI assistant and human annotations, we employ this template to double-check. It filters out samples with incorrect answers and reasoning processes.}
    \label{fig:prompts_data_filtering}
\end{figure*}

\begin{figure*}[t]
    \centering
    \includegraphics[width=\textwidth]{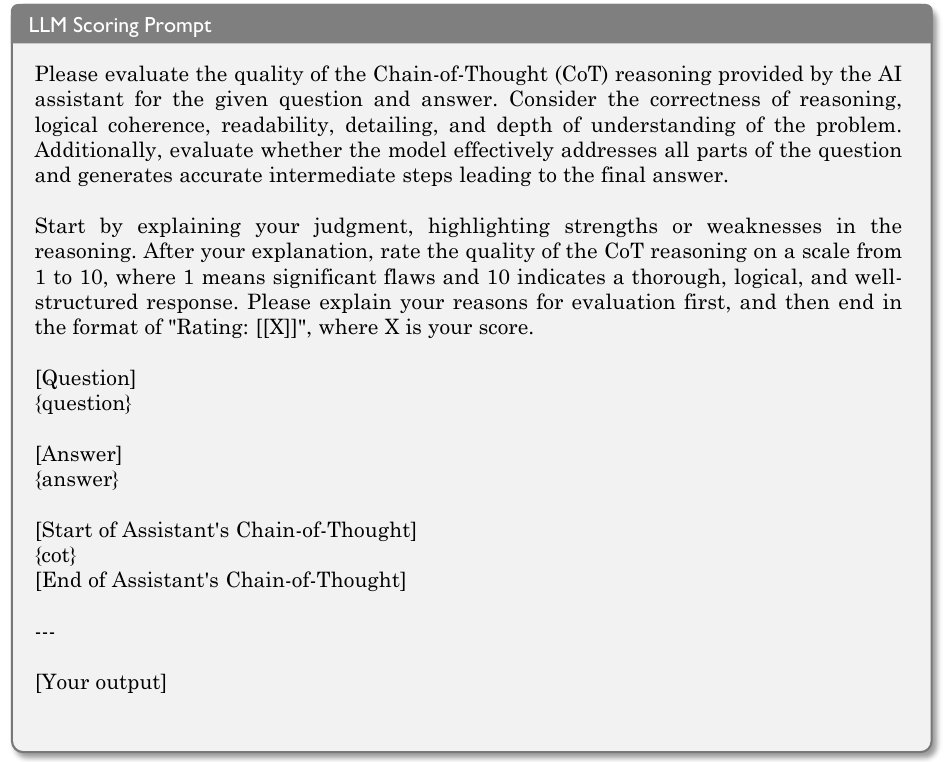} 
    \caption{Prompt for GPT scoring. We use this template and GPT-4o to quantitatively evaluate the quality of the generated data. The results show that our AMATH data outperforms human annotations in terms of AI preference scores.}
    \label{fig:prompts_scoring}
\end{figure*}

In this section, we present the prompt used in self-structured CoT~\ref{fig:prompts_atomthink} and multimodal CoT annotation engine. Prompts in data engine include: long CoT generation (Figure \ref{fig:prompts_dynamic}), short CoT augmentation (Figure \ref{fig:prompts_cot_aug}), data filtering (Figure \ref{fig:prompts_data_filtering}), and quality scoring (Figure \ref{fig:prompts_scoring}).


\begin{figure*}[t]
    \centering
    \includegraphics[width=\textwidth]{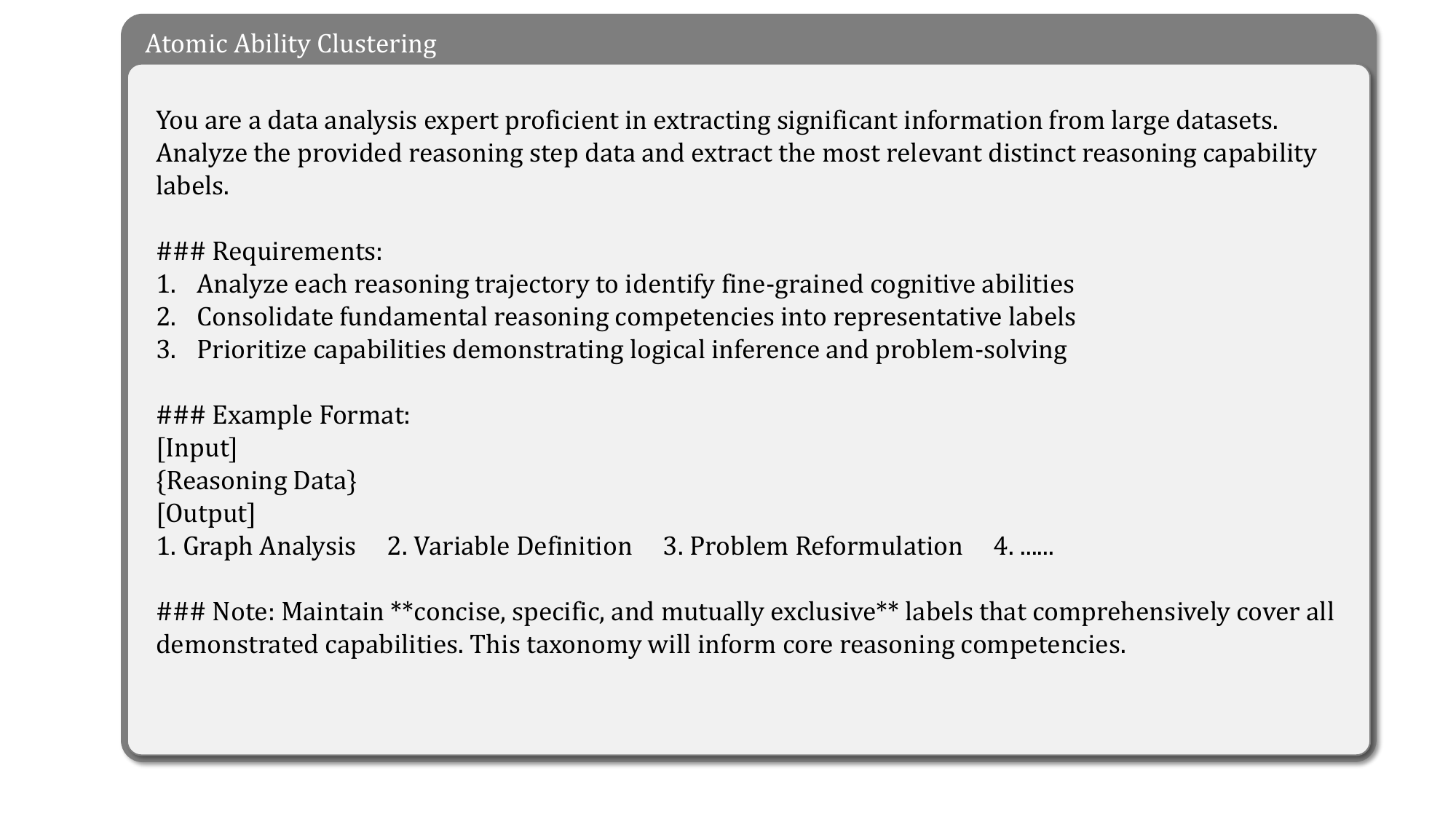} 
    \caption{Prompt for clustering the reasoning behaviors in GPT-4o.}
    \label{fig:atomic_ability}
\end{figure*}

\section{Cases of AtomThink output}
\begin{figure*}[t]
    \centering
    \includegraphics[width=\textwidth]{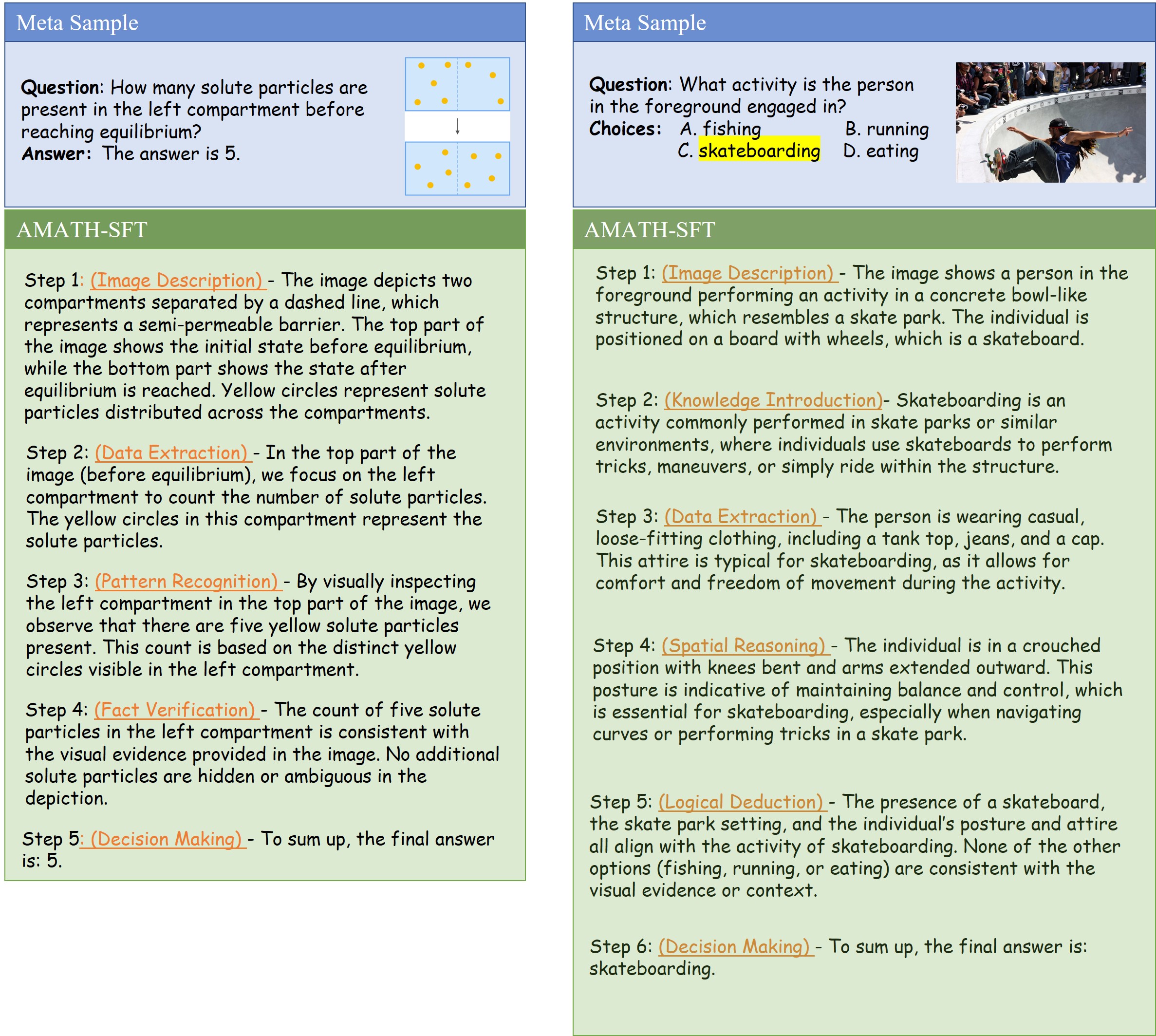} 
    \caption{Case of AtomThink-LlamaV output. Based on the type of problem and information provided, model autonomously explores diverse reasoning behaviors, e.g. Fact Verification, Spatial Reasoning and Logical Deduction.}
    \label{fig:case_llamav1}
\end{figure*}

\begin{figure*}[t]
    \centering
    \includegraphics[width=\textwidth]{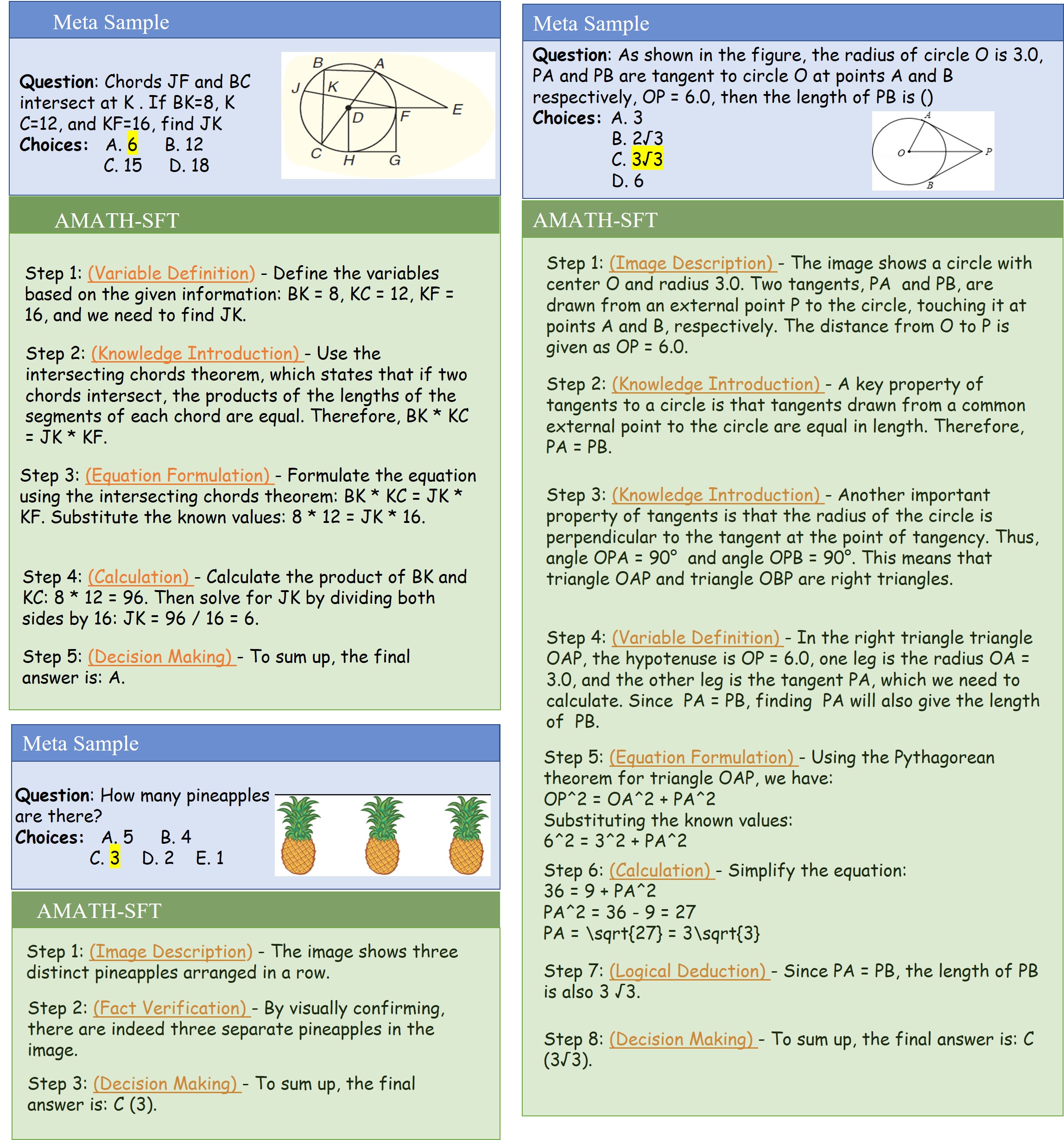} 
    \caption{Case of AtomThink-LlamaV output. Depending on the difficulty of problem, it generates shorter or longer CoTs without human supervision.}
    \label{fig:case_llamav2}
\end{figure*}

In Figure \ref{fig:case_llamav1} and Figure \ref{fig:case_llamav2}, we present the SCoT outputs generated by Llama3.2-Vision-11B models trained with AtomThink. Compared to original models, we are able to produce a dynamic thinking process similar to OpenAI-o1~\cite{openai2024_o1}. The models tend to use image features to decompose challenging mathematical problems into multiple atomic sub-questions, which are then solved step by step. It is capable of generating responses of varying lengths based on the difficulty of problem and exhibits diverse reasoning behaviors (including Fact Verification, Spatial Reasoning and Logical Deduction). The results demonstrate that our outputs are more accurate in recognizing visual information and reduce reasoning hallucinations.

\section{Cases of AtomMATH Dataset}
\begin{figure*}[t]
    \centering
    \includegraphics[width=\textwidth]{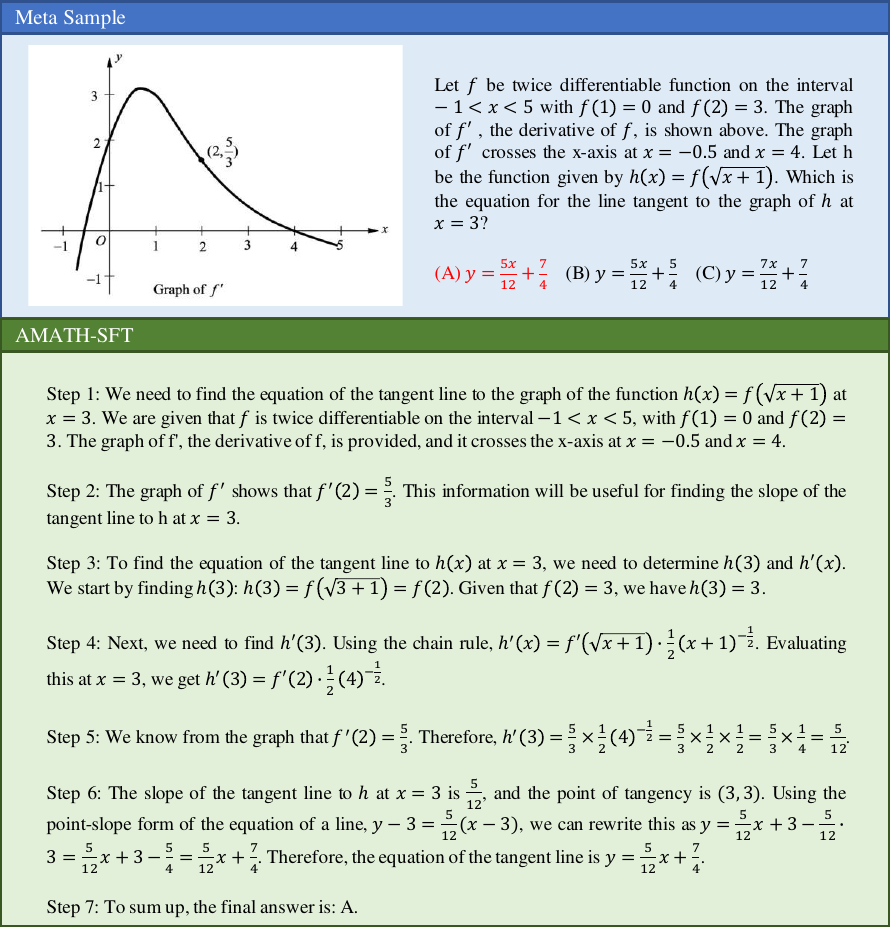} 
    \caption{Case of AMATH-SFT.}
    \label{fig:case_sft1}
\end{figure*}

\begin{figure*}[t]
    \centering
    \includegraphics[width=\textwidth]{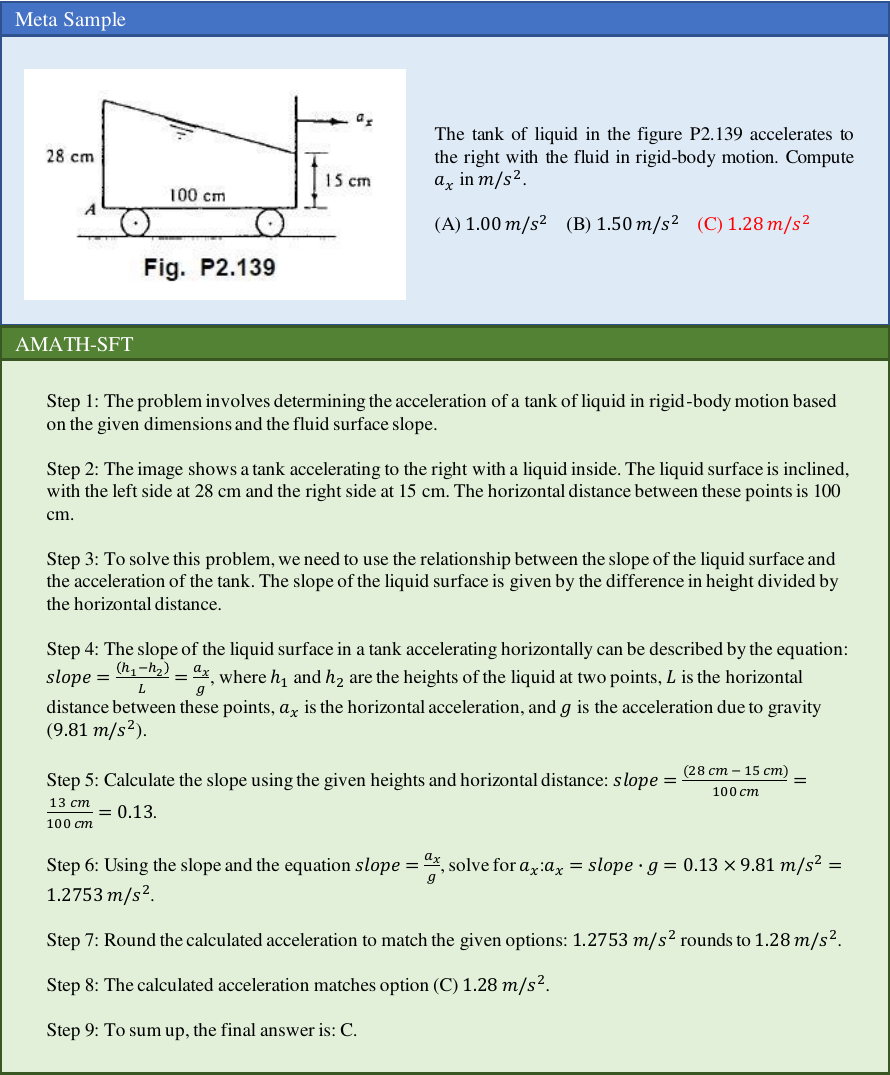} 
    \caption{Case of AMATH-SFT.}
    \label{fig:case_sft2}
\end{figure*}

To further illustrate the AMATH dataset we released, we randomly select several examples. Figure \ref{fig:case_sft1} and Figure \ref{fig:case_sft2} show two case from AMATH-SFT.

\begin{table*}[h!]
\centering
\begin{tabular}{
    c 
    c 
    |
    c 
    c 
    c 
    |
    c 
    c 
    c 
    c 
    c 
    c 
}
\toprule
& & \multicolumn{3}{c|}{\textbf{MathVista}} & \multicolumn{6}{c}{\textbf{MathVerse}} \\
\midrule
\textbf{Model} & \textbf{Inference} & \textbf{General} & \textbf{Math} & \textbf{Total} & \textbf{TL} & \textbf{TD} & \textbf{VI} & \textbf{VD} & \textbf{VO} & \textbf{Total} \\
\midrule
LLaVA-Llama3-8B & Direct & 34.1 & 25.6 & 29.5 & 16.0 & 19.3 & 16.4 & 13.1 & 15.0 & 15.9\\
LLaVA w/. Formatted & CoT & 30.2 & 22.9 & 26.3 & 14.3 & 18.4 & 15.7 & 10.0 & 7.7 &13.2 \\
AtomThink-LLaVA-Llama3 & Direct & 34.4 & 27.2 & 30.5 & 16.0 & 19.3 & 16.2 & 13.1 & 15.0 & 15.9\\
AtomThink-LLaVA-Llama3 & SCoT & \textbf{36.9} & \textbf{37.0} & \textbf{36.6} & \textbf{22.2} & \textbf{26.6} & \textbf{24.1} & \textbf{20.9} & \textbf{17.9} & \textbf{22.4}\\
AtomThink-LLaVA-Llama3 & SCoT w./ PRM & \textbf{36.5} & \textbf{41.3} & \textbf{39.1} & \textbf{36.1} & \textbf{42.4} &\textbf{30.0}  &\textbf{36.8}  & \textbf{28.6 }&\textbf{34.7} \\
\midrule
EMOVA-8B-200k & Direct & 52.4 & 51.1 & 51.7 & 34.4 & 39.0 & 33.4 & 30.1 & 23.5 & 32.1 \\
EMOVA w/. Formatted & CoT & 30.9 & 31.3 & 31.1 & 26.5 & 36.5 & 25.3 & 20.4 & 19.8 & 25.7 \\
AtomThink-EMOVA & Direct & 53.9 & 52.4 & 53.1 & 33.6 & 39.0 & 33.8 & 28.0 & 24.4 & 31.8\\
AtomThink-EMOVA & SCoT & 48.7 & \textbf{54.4} & \textbf{51.8} & \textbf{36.5} & \textbf{42.4} & \textbf{34.1} & \textbf{32.9} & \textbf{29.7} & \textbf{35.1} \\
AtomThink-EMOVA & SCoT w./ PRM & 48.9 & \textbf{57.0} & \textbf{53.3} &\textbf{42.1}& \textbf{51.5}&\textbf{39.0} &\textbf{36.7} &\textbf{33.1} & \textbf{40.5} \\

\bottomrule
\end{tabular}
\caption{Comparison of accuracy on MathVista and MathVerse. Our AtomThink-LLaVA-Llama3 outperforms the baseline in all sub-tasks across two benchmarks, achieving an average improvement of 14.2\%.}
\label{tab:app_main}
\end{table*}

\begin{figure}[ht]
    \centering
    \hspace{-0.5cm}
    \includegraphics[width=0.5\textwidth]{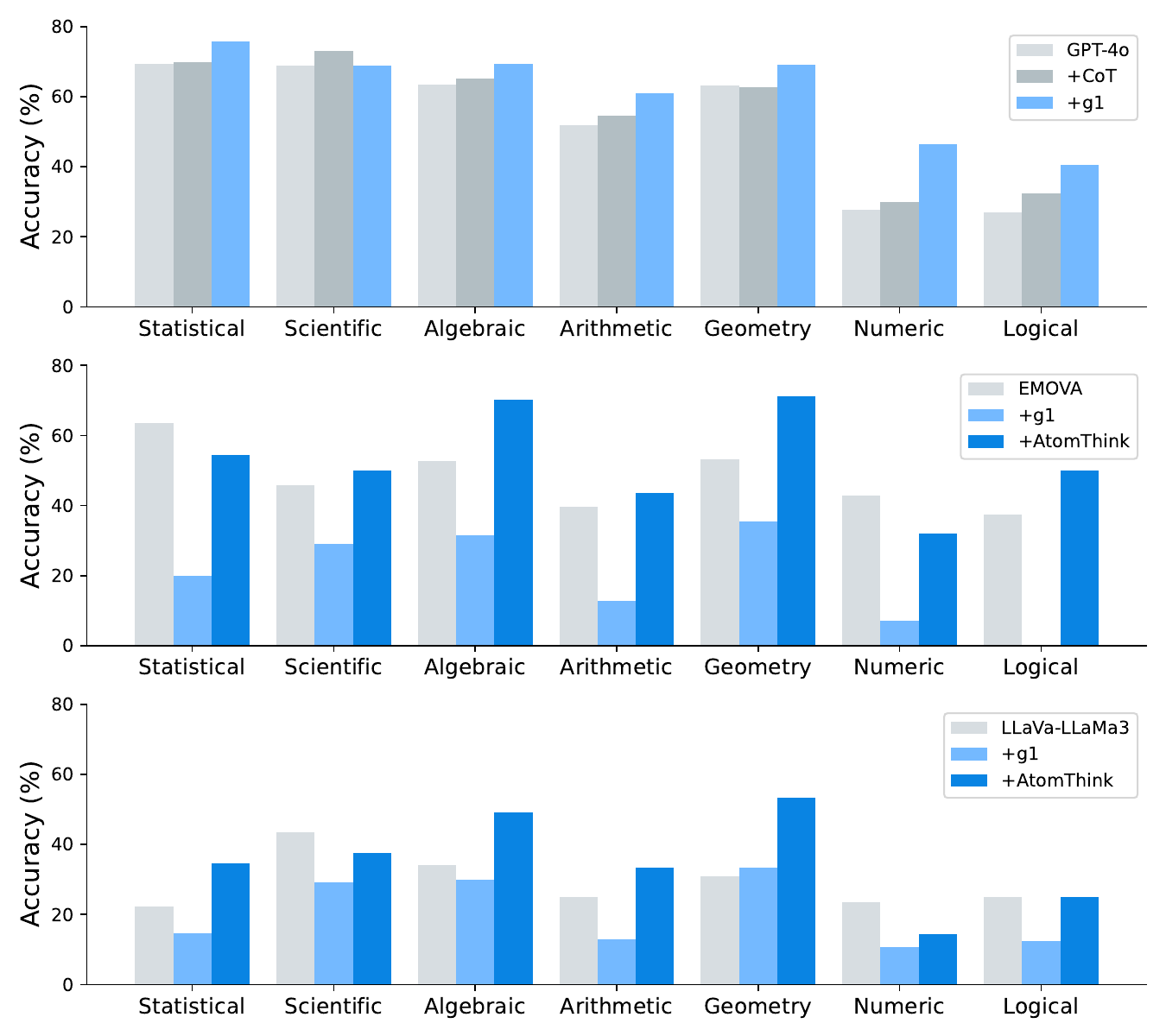} 
    \caption{Comparison to CoT and g1 in MathVista subsets. In contrast to the declining trend observed in g1, AtomThink outperforms the baseline across most subsets.}
    \label{fig:g1}
\end{figure}

\begin{table}[ht]
\centering
\resizebox{\columnwidth}{!}{ 
\begin{tabular}{
    l 
    l 
    |
    c 
    c
    c 
    c 
    c 
}
\toprule
\textbf{Model} & \textbf{Method} & \textbf{MathVista-M} & \textbf{MathVista-G} & \textbf{MathVerse} \\
\midrule
EMOVA-200k & Direct & 51.1 & 52.4 & 33.3 & \\
\midrule
\multicolumn{1}{l}{\multirow{2}*{AtomThink}} & Direct & 52.4 & 53.9 & 35.7 & \\
 & Quick Think & 54.2 & 46.7 &38.0 &\\
\midrule
\multicolumn{1}{l}{\multirow{4}*{w/. Path-wise}}  & Majority Voting & 48.8 & 49.4 & 39.0 & \\
 & BoN-Last & 51.2 & 46.8 & 41.3 & \\
 & BoN-Avg & 58.7 & 40.5 & 38.7 \\
 & BoN-Min & 53.7 & 53.2 & 40.0 \\
 \midrule
\multicolumn{1}{l}{\multirow{2}*{w/. Step-wise}} & Greedy & 46.3 & 45.6 & 38.3 & \\
 & Beam Search & 57.1 & 53.2 & 45.3 \\
\bottomrule
\end{tabular}
}
\caption{Ablation study on Path-wise and step-wise search. The results show that both Best-of-N-Min(BoN-Min) and Beam Search exhibit consistent performance improvements.}
\label{tab:search}
\end{table}

\begin{table}[ht]
    \centering
    \begin{tabular}{ccc}
        \toprule
        Candidate & AtomThink-EMOVA & AtomThink-LLaVA \\
        \midrule
        1 & 38.0 & 22.5 \\
        2 & 37.5 & 23.0 \\
        3 & 38.5 & 31.5 \\
        4 & 41.5 & 33.5 \\
        5 & 42.5 & 35.5 \\
        6 & 45.5 & 36.0 \\
        10 & 45.5 & 38.0 \\
        \bottomrule
    \end{tabular}
    \caption{Test-time scaling law of MLLMs. Results suggest that increasing the number of candidates during inference can enhance models' performance on mathematical tasks, aligning with the finding of test-time scaling laws in language models.}
    \label{tab:scaling_law}
\end{table}

\section{Early experiments}
In this section, we provide more early exploration in SFT based slow thinking model. We use LLaVA-Llama3~\cite{liu2024visual} and EMOVA-8B~\cite{chen2024emova} to perform experiments.

\subsection{Implementation Details}

\paragraph{Basic Model.} For LLaVA-Llama3~\cite{liu2024visual}, we choose the pre-trained ViT-L/14 of CLIP~\cite{radford2021learning} as the vision encoder and Llama3-8B~\cite{dubey2024llama} as our LLM. To align visual features with the LLM, we incorporates a Multi-Layer Perceptron (MLP) as a projector between the visual encoder and the language model. For EMOVA-8B~\cite{chen2024emova}, we use the original setting of EMOVA that uses InternViT-6B~\cite{chen2024internvl} and LLaMA-3.1-8B~\cite{dubey2024llama}. The C-Abstractor~\cite{cha2024honeybee} with two ResBlocks is adopted as the projector.

The training of LLaVA-Llama-3-8B follows a structured two-stage process~\cite{liu2024visual}. In our experiment, we only load its weights from pre-training stage and deploy supervised fine-tuning. During SFT, the training data comprises the LLaVA-Instruct-665k, a 46k subset of PRM800k and our AMATH-SFT dataset. The weights of language model and MLP projector are unfreezed. The model undergoes an epoch of training with a reduced learning rate of 2e-5 and batch size of 128. To create AtomThink-EMOVA, we post-train EMOVA using AMATH-SFT and a sub-sampled dataset of EMOVA-SFT-4m, containing 200k samples. During this process, the weights of the LLM and the C-Abstractor projector are updated. EMOVA is fine-tuned for 1 epoch with a batch size of 128 and a learning rate of 2e-6.

\paragraph{PRM Setting.} We initially fine-tuned a large language model to introduce textual process supervision. We utilize the pre-trained Qwen2.5-Math-7B-Instruct~\cite{yang2024qwen2} and Math-psa-7B~\cite{wang2024openr} models as our foundational architectures. Qwen2.5-Math-7B-Instruct is an open-source large language model specifically designed for mathematical reasoning tasks. Math-psa-7B is a text-based process supervision model trained using datasets such as PRM800K~\cite{lightman2023let}, Math-Shepherd~\cite{Math-shepherd} and MATH-APS~\cite{wang2024openr}. Low-Rank Adaptation (LoRA) is applied to fine-tune with the following parameters: rank (r) of 8, alpha scaling factor of 32, dropout rate of 0.1, and targeting the q and v projectors. Training is conducted over one epoch with a batch size of 256 and a learning rate of 1e-5. We sample a 20k-instance training set from PRM800K and combine it with the AMATH-PRM dataset, which is derived from multimodal CoT annotations, to serve as our fine-tuning data. All the samples include question, historical steps, and current step, with each current step being assigned a label of either correct or incorrect. In line with OpenR's settings, we designate "\textbackslash n\textbackslash n\textbackslash n\textbackslash n\textbackslash n" as the step separator and return the conditional probability of the current step being correct.

\subsection{Results}

\paragraph{Multimodal Performance Improvement.} In Table~\ref{tab:app_main}, our AtomThink framework is applied to train LLaVA-Llama3-8B and EMOVA-8B, yielding consistent performance improvements over the original models. When combined with PRM, AtomThink-EMOVA achieves a new state-of-the-art on MathVerse and narrowing the gap between MLLMs and human performance. On MathVista, it also achieves performance to 53.3\%. These results demonstrate the framework's strong generalization capability. In Figure \ref{fig:g1}, we compare AtomThink with the state-of-the-art open-source inference strategy, g1\footnote{https://github.com/bklieger-groq/g1}, which employs dynamic prompting to make model focus on single step reflection. In GPT-4o, direct application of g1 for multi-turn reasoning yields a greater improvement over Chain-of-Thought, particularly in numeric and geometric tasks. However, due to the reliance on the inherent reasoning capabilities of large-scale language models, its performance significantly degrades on smaller models such as EMOVA-8B and LLaVA-Llama3-8B. In contrast, our AtomThink framework consistently enhances the performance of these MLLMs.

\paragraph{Test-time scaling law in multimodal tasks}
In this subsection, we evaluate the impact of inference-time expansion on experimental outcomes using a 200-sample subset of the MathVerse dataset. We employ the AtomThink-EMOVA-8B and AtomThink-LLaVA-8B in Sec.~\ref{sec:exp} as base models, maintaining a fixed beam size of 1, and increase the number of candidates per sampling to extend test time. The results in Table \ref{tab:scaling_law} indicate that, as the number of candidates increases, both models exhibit consistent performance improvements on multimodal mathematical tasks. Specifically, AtomThink-EMOVA-8B's accuracy rises from 38.0\% with a single candidate to 45.5\% with 6 candidates, while AtomThink-LLaVA-8B's accuracy improves from 22.5\% to 36.0\% over the same range.

\paragraph{Evaluate CoT Capability with Various Prompts}

Chain-of-Thought prompting has been demonstrated to enhance the performance of large language models by guiding them to deliberate on their predictions~\cite{wei2022chain}. However, previous studies have indicated that such benefits emerge predominantly in models exceeding 100 billion parameters. In our investigation, we extend this analysis to multimodal large language models by employing various widely-used prompts to induce step-by-step reasoning in the 8B-parameter LLaVA-Llama3 model. In Table \ref{tab:cot_prompt}, the prompt "Answer the question using a single word or phrase." forces model output directly and achieves the highest accuracy at 18.5\%. In contrast, prompts explicitly instructing step-by-step reasoning, such as "Let's think step by step." results in lower accuracies of 11.1\%. Other prompts that encourage CoT output also lead to a significant decrease in reasoning accuracy. These findings suggest that incorporating slow-thinking capabilities into smaller models presents substantial challenges.

\begin{table*}[htbp]
    \centering
    \begin{tabular}{cc}
        \toprule
        Prompt & MathVerse200 \\
        \midrule
        Answer the question using a single word or phrase. & 18.5 \\
        \midrule
        Let's think step by step. & 11.0 \\
        \midrule
        First perform reasoning, then finally answer the question and &\multirow{2}{*}{9.5} \\ provide the final value at the end in the following format: Answer: xxx. &  \\
        \midrule
        Answer the following question step by step concisely. & 14.5 \\
        \midrule
       Given the following problem, reason and give a final answer to the problem. Your response &\multirow{2}{*}{15.5} \\ should end with "The final answer is [answer]" where [answer] is the response to the problem. &   \\
        \midrule
        Answer the following question. The last line of your response should be of the following &\multirow{2}{*}{14.0}\\format: 'Answer: xxx' where 'xxx' is the answer. Think step by step before answering.&  \\
        \bottomrule
    \end{tabular}
    \caption{Evaluate the CoT performance of LLaVA-Llama3-8B using different prompts. Results indicate that smaller multimodal large language models exhibit limited CoT capabilities.}
    \label{tab:cot_prompt}

\end{table*}

\section{Limitations}
Due to a lack of computational resources, we did not conduct research on larger-scale visual models. Additionally, although we performed manual sampling and screening during data creation, it may not cover all cases and atomic steps.

\end{document}